\documentclass[sigconf, nonacm=true]{acmart}

\AtBeginDocument{%
  \providecommand\BibTeX{{%
    \normalfont B\kern-0.5em{\scshape i\kern-0.25em b}\kern-0.8em\TeX}}}

\usepackage{algorithm}
\usepackage{algorithmicx}
\usepackage{algpseudocode}
\usepackage{amsthm}
\usepackage{subfigure}
\usepackage{threeparttable}

\newtheorem{Def}{Definition}
\newtheorem{The}{Theorem}
\newtheorem{Rem}{Remark}
\newtheorem*{OP}{Output Perturbation}
\newtheorem*{ObP}{Objective Perturbation}
\newtheorem*{GP}{Gradient Perturbation}
\newtheorem*{C1}{A Bridge between Local and Central Differenital Privacy}
\newtheorem*{C2}{Superior Theoretical and Experimental Results}
\newtheorem*{C3}{A More General Condition}

\begin{document}

\title{Input Perturbation: A New Paradigm between Central and Local Differential Privacy}

\author{Yilin Kang}
\email{kangyilin@iie.ac.cn}
\affiliation{
	\institution{Institute of Information Engineering, Chinese Academy of Sciences}
}
\author{Yong Liu}
\email{liuyong@iie.ac.cn}
\affiliation{
	\institution{Institute of Information Engineering, Chinese Academy of Sciences}
}
\author{Ben Niu}
\email{niuben@iie.ac.cn}
\affiliation{
	\institution{Institute of Information Engineering, Chinese Academy of Sciences}
}
\author{Xinyi Tong}
\email{tongxinyi@buaa.edu.cn}
\affiliation{
	\institution{Beihang University}
}
\author{Likun Zhang}
\email{2016312357@email.cufe.edu.cn}
\affiliation{
	\institution{Central University of Finance and Economics}
}
\author{Weiping Wang}
\email{wangweiping@iie.ac.cn}
\affiliation{
	\institution{Institute of Information Engineering, Chinese Academy of Sciences}
}
%
%
%
%
%
%
%

\renewcommand{\shortauthors}{Kang et al.}

\begin{abstract}
  Traditionally, there are two models on differential privacy: the central model and the local model.
  The central model focuses on the machine learning model and the local model focuses on the training data.
  In this paper, we study the \textit{input perturbation} method in differentially private empirical risk minimization (DP-ERM), preserving privacy of the central model.
  By adding noise to the original training data and training with the `perturbed data', we achieve ($\epsilon$,$\delta$)-differential privacy on the final model, along with some kind of privacy on the original data.
  We observe that there is an interesting connection between the local model and the central model: the perturbation on the original data causes the perturbation on the gradient, and finally the model parameters.
  This observation means that our method builds a bridge between local and central model, protecting the data, the gradient and the model simultaneously, which is more superior than previous central methods.
  Detailed theoretical analysis and experiments show that our method achieves almost the same (or even better) performance as some of the best previous central methods with more protections on privacy, which is an attractive result.
  Moreover, we extend our method to a more general case: the loss function satisfies the Polyak-Lojasiewicz condition, which is more general than strong convexity, the constraint on the loss function in most previous work.
\end{abstract}



\keywords{differential privacy, machine learning, input perturbation, empirical risk minimization}


\maketitle

\section{Introduction}
In recent years, machine learning has been shown effective in fields such as pattern recognition and data mining \cite{1}, \cite{2}, \cite{3}, \cite{4} and large quantities of personal data has been collected to support machine learning algorithms.
The collection of tremendous data leads a huge problem: the disclosure of personal sensitive information.
In real scenarios, not only the leakage of original data will disclose the information of individuals, when training machine learning models, model parameters may reveal sensitive information in an undirect way as well \cite{6}, \cite{5}.

To solve the problem of information leakage, differential privacy (DP) \cite{7}, \cite{8} was proposed and has become a popular way to preserve privacy in machine learning.
It preserves sensitive information by adding random noise, making an adversary can not infer any single data instance in the dataset by observing model parameters.
Differential privacy has received a great deal of attentions and has been applied to regression \cite{9}, \cite{10}, \cite{51}, boosting \cite{16}, \cite{17}, PCA \cite{11}, \cite{12}, GAN \cite{24}, \cite{36}, transfer learning \cite{13}, graph algorithms \cite{25}, \cite{52}, \cite{56}, deep learning \cite{14}, \cite{15} and other fields.

Empirical risk minimization (ERM), covering a wide variety of machine learning tasks, is also bothered by privacy problems.
There is a long list of works on DP-ERM \cite{20}, \cite{21}, \cite{18}, \cite{22}, \cite{23}.
According to different ways of adding noise, three approaches were proposed to achieve differential privacy: output perturbation, objective perturbation  and gradient perturbation, adding noise to the final model, the objective function and the gradient, respectively.

However, the original data is not preserved by perturbation methods mentioned above.
In real scenarios, before training, original data is sent to a `data center', which is trusted in central models, shown in Figure 1 (a).
When it comes to the situation that `data center' is not trusted, local differential privacy (LDP) \cite{45}, \cite{40} was proposed to provide \textit{plausible deniability}, by randomizing the data before releasing it.
As shown in Figure 1 (b), LDP focuses on the privacy of the communications between individuals and the `server', rather than the final machine learning model \cite{48}, \cite{43}, \cite{46}, \cite{59}, \cite{41}.
However, the noise added for preserving privacy in LDP is always large, compromising predictive performance.

\begin{figure}[htbp]
\centering
\subfigure[Central differential privacy]{\includegraphics[width=0.4\textwidth]{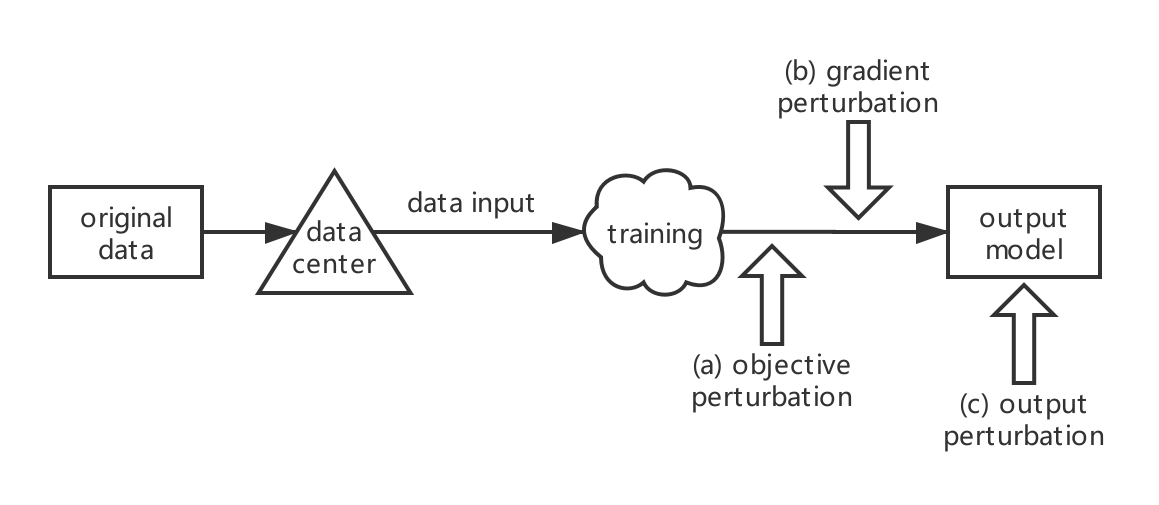}}
\subfigure[Local differential privacy]{\includegraphics[width=0.4\textwidth]{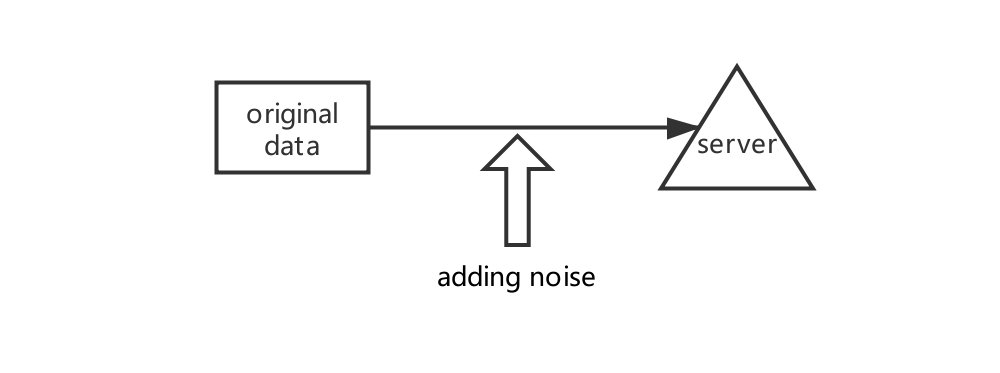}}
\subfigure[Our Method]{\includegraphics[width=0.4\textwidth]{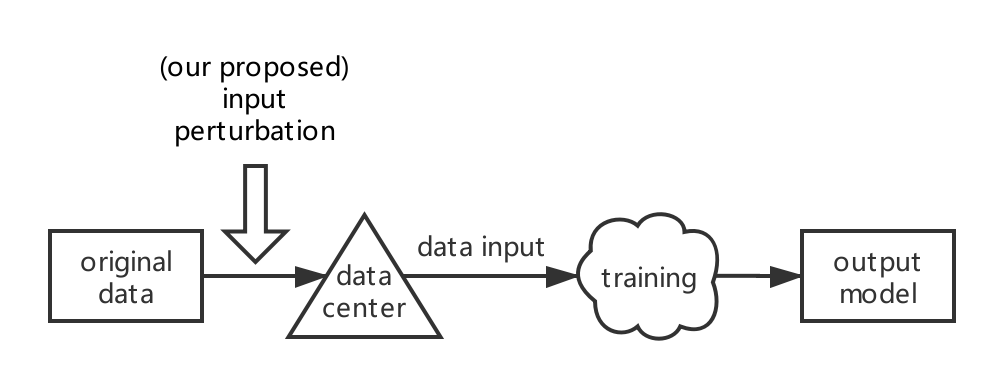}}
\caption{Different perturbation methods.}
\end{figure}

To alleviate the problems mentioned above, in this paper, we study the \textit{input perturbation} method, achieving ($\epsilon$,$\delta$)-differential privacy on the final model.
The comparison between our method and previous perturbation methods is shown in Figure 1.
It can be observed that our method focuses on the final model and preserves the original data to some extents.
Even if the adversaries get the perturbed data in the `data center', the leakage of sensitive information decreases a lot compared with traditional central models.
Actually, adding noise to original data to preserve privacy is commonly used in the field of computer vision \cite{58}, \cite{44}, \cite{63}.
In this way, it is not easy to reconstruct the original data \cite{49}.

By adding noise to original data, protections are applied before `data input', and our method is more reliable than traditional central models.
Moreover, we observe that our input perturbation method also perturbs the gradient and the final model parameters, building a bridge between local and central differential privacy.

\subsection*{Contributions of Our Method}
\begin{C1}
\rm{
By observing a fact that the noise added to data causes perturbation on the gradient and finally the final model, we build a bridge between local and central differential privacy, guaranteeing ($\epsilon$,$\delta$)-differential privacy on the final model along with some kind of privacy on the original data simultaneously.
When comparing with traditional central perturbation methods, in which the privacy of original data is ignored, we provide more privacy.
Meanwhile, comparing with LDP, we make a balance on the performance and the privacy of individuals: adding less noise and keeping better performance.
Additionally, the privacy on the final model remains.
}
\end{C1}

\begin{C2}
\rm{
Detailed theoretical analysis and experiments show that the performance of our method is similar to (or even better than) some of the best previous methods in central setting.
Considering that our method preserves both the original data and the final model and other central methods ignore the security of original training data, the results are attractive.
When it comes to LDP, although in our method, the privacy between individuals and the `data center' is weaker, the performance of our method is much better, which is a trade-off and the sacrifice is acceptable.
}
\end{C2}

\begin{table*}
\begin{center}
{\caption{Comparison between our method and other methods on noise bound and excess empirical risk bound.}}
\begin{threeparttable}
\begin{tabular}{c|c|c|c|c|c}
\hline
\rule{0pt}{6pt}
 & Method & $\delta=0$ & Noise Type & Noise Bound & Excess Empirical Risk Bound \\
\hline
\cite{18} & Output Perturbation & Yes & $v(b)=\frac{1}{\alpha}e^{-\gamma\|b\|}$ & $\gamma=O\left(n\lambda\epsilon\right)$ & $O\left(\frac{p^2\log^2(p/\delta)(L+\lambda)}{\lambda^2n^2\epsilon^2}\right)$ \\
\cite{18} & Objective Perturbation & Yes & $v(b)=\frac{1}{\alpha}e^{-\gamma\|b\|}$ & $\gamma=O\left(\epsilon-\log(1+\frac{2L}{n\lambda}+\frac{L^2}{n^2\lambda^2})\right)$ & $O\left(\frac{p^2\log^2(p/\delta)}{\lambda n^2\epsilon^2}\right)$ \\
\cite{23} & Objective Perturbation & No & Gaussian Noise\tnote{2} & $O\left(\frac{\zeta^2(\epsilon+\log(4/\delta^2))}{\epsilon^2}\right)$ & $O\left(\frac{\|b\|^2}{n^2\Delta}+\Delta\|\hat{\theta}\|^2\right)$ \\
\cite{21} & Gradient Perturbation & No & Gaussian Noise & $O\left(\frac{G^2n^2\log(n/\delta)\log(1/\delta)}{\epsilon^2}\right)$ & $O\left(\frac{G^2p\log^2(n/\delta)\log(1/\delta)}{n\Delta\epsilon^2}\right)$ \\
\cite{22} & Output Perturbation & No & Gaussian Noise & $O\left(\frac{G^2(1+L/\Delta)^2\log(2/\delta)}{n^2L^2\epsilon^2}\right)$ & $O\left(\frac{L G^2p\log(1/\delta)}{n^2\epsilon^2\Delta^2}\right)$ \\
\cite{20} (DP-SVRG) & Gradient Perturbation & No & Gaussian Noise & $O\left(\frac{G^2Tm\log(1/\delta)}{n^2\epsilon^2}\right)$ & $O\left(\frac{G^2p\log(n)\log(1/\delta)}{n^2\Delta\epsilon^2}\right)$ \\
\cite{20} (traditional) & Gradient Perturbation & No & Gaussian Noise & $O\left(\frac{G^2T\log(1/\delta)}{n^2\epsilon^2}\right)$ & $O\left(\frac{G^2p\log(n)\log(1/\delta)}{n^2\epsilon^2}\right)$ \\
\cite{48} & LDP & Yes & Randomized response & None\tnote{3} & $O\left(\frac{Gp}{\epsilon\sqrt{n}}\right)$ \\
\cite{43} & LDP & Yes & Laplace Noise\tnote{2} & $O\left(pn\epsilon^2\right)$ & $O\left(\frac{\sqrt{p}}{(\sqrt{n}\epsilon)^{\frac{L}{L+2p}}}\right)$ \\
\cite{62} & Input Perturbation & No & Gaussian Noise & $O\left(\frac{G^2(\log(16/\delta)+\epsilon)}{n\epsilon^2}\right)$ & $O\left(\frac{pG^2(\log(16/\delta^2)+\epsilon)}{Ln\epsilon}\right)$ \\
\hline
Our Method & Input Perturbation & No & Gaussian Noise & $O\left(\frac{G^2T\log(1/\delta)}{n(n-1)\sqrt{\Delta}\epsilon^2}\right)$ & $O\left(\frac{\alpha(2LD+G)G^3d\log^2(n)\log(1/\delta)}{n(n-1)\sqrt{\Delta}\epsilon^2}\right)$ \\
\hline
\end{tabular}
\begin{tablenotes}
\item[2] The noise bound of the Gaussian and Laplace noise and are represented by the variance, whose means are 0.
\item[3] The noise added by randomized response is complicated, details can be found in \cite{48}.
\item[4] $n$ is the size of training set, $T$ is the number of total iterations, $p$ is the number of model parameters, input $x$ has $d$-dimensional feature.
\end{tablenotes}
\end{threeparttable}
\end{center}
\end{table*}

\begin{C3}
\rm{
Considering that most previous works assume the loss function is strongly convex, we generalize it to the condition that the loss function satisfies the Polyak-Lojasiewicz condition, which is more general than strong convexity.
}
\end{C3}

The rest of the paper is organized as follows.
In Section 2, we introduce some works related to our method.
We introduce some basic definitions and formulations in Section 3.
In Section 4, we propose our method: \textit{input perturbation} in detail.
In Section 5, we give the theoretical analysis of our method and extend it to a more general case.
We present the experimental results in Section 6.
Finally, we conclude the paper in Section 7.

\section{RELATED WORK}
In this section, we introduce some work on private ERM methods and list the comparison of their theoretical results.

The first work on DP-ERM was proposed in \cite{18}, in which two methods were proposed: output perturbation and objective perturbation.
The probability density function of the noise $v(b)=\frac{1}{\alpha}e^{-\gamma\|b\|}$, where $\alpha$ is a normalizing constant, $\gamma$ is a function of the privacy budget $\epsilon$ and $\|\cdot\|$ denotes $\ell_2$-norm.
In this work, the derivative of the loss function $\nabla\ell(\cdot)$ was assumed $L$-Lipschitz.
Based on these assumptions, it provided theoretical analysis on the noise bound and the excess empirical risk bound.
The noise of the method proposed in \cite{18} was improved by \cite{23}.
The improved noise is related to the upper bound of $\|\nabla\ell(\cdot)\|$, $\zeta$ (i.e. $\|\nabla\ell(\theta)\| \leq \zeta$ for all $\theta$).
Additionally, this work assumed the perturbed objective function is $\Delta$-strongly convex, and gives the excess empirical risk bound, which is related to the noise $b$ and the optimal model $\hat{\theta}$.

By gradient perturbation, \cite{21} added noise to the gradient, guaranteeing differential privacy by assuming that the loss function $\ell(\cdot)$ is $G$-Lipschitz.
Like in \cite{18}, \cite{22} proposed an output perturbation method, achieving a better excess empirical bound.
Advanced gradient descent method Prox-SVRG \cite{26} was introduced in \cite{20}, and a new algorithm DP-SVRG was proposed.
DP-SVRG achieved optimal or near optimal utility bounds with less gradient complexity.
In this work, the noise bound was related to $m$, the sampling iterations in the algorithm DP-SVRG.
Note that in DP-SVRG, better results are because of advanced gradient descent method, rather than advanced perturbation method.

However, all the methods proposed in previous work are based on output perturbation, objective perturbation or gradient perturbation.
As a result, privacy preserving is after `data input', which increases the risk of information leakage.
Although LDP can solve the problem of `untrusted data center', the theoretical results are much worse, which can be observed in Table 1\footnote[1]{The theoretical results of LDP and input perturbation in Table 1 are simplified, more details can be found in \cite{43}, \cite{48} and \cite{62}.}.

Under these circumstances, input perturbation was proposed in \cite{62}, in which although noise is added to data, it achieves differential privacy by constructing a `perturbed objective function'.
It guarantees ($O(\sqrt{n\epsilon})$,$\delta$)-LDP and ($\epsilon$,$\delta$)-central DP.
However, considering that $n$ is always large, the LDP is unsatisfactory.
Moreover, its excess empirical risk bound is also much weaker than some central models because the noise added to the original data is large.

Considering the problems mentioned above, in this paper, we focus on \textit{input perturbation}, adding noise to the original data and training machine learning model by the `perturbed data'.
By observing the effects caused by input perturbation: noise added to the data leads perturbation on the gradient and the final model parameters, our method provides ($\epsilon$,$\delta$)-differential privacy on the final model, which is the same as central setting, along with some kinds of protections on original data, showing the connections between local and central differential privacy.
Theoretical comparisons between our method and previous methods are shown in Table 1.

It can be observed that the noise bound of our method is better than the gradient perturbation method proposed in \cite{20}.
For which the advanced gradient descent algorithm DP-SVRG is used, the difference is by a factor of $m\sqrt{\Delta}$.
For traditional gradient descent method in \cite{20}, the difference is $\sqrt{\Delta}$.
When comparing with the method proposed in \cite{21}, our method is much better, the difference is up to $\frac{n^4\sqrt{\Delta}\log(n/\delta)}{T}$.
When it comes to the input perturbation method proposed in \cite{62}, our noise bound is better than it approximately by a factor of $n$.

The excess empirical risk bound of our method is related to the upper bound of the $\ell_2$-norm of the model parameters, $D$ (i.e. $\|\theta\| \leq D$).
Our method is better than traditional gradient perturbation method proposed in \cite{21} by a factor of $\frac{\alpha(2LD+G)Gd\sqrt{\Delta}}{np}$, almost $\frac{d}{np}$, considering $\alpha(2LD+G)G\sqrt{\Delta}$ can be seemed as a constant.
When comparing with the methods proposed in \cite{20}, our method achieves almost the same excess empirical risk bound, the difference is approximately $\frac{d\log(n)}{p}$, no matter the advanced gradient descent algorithm, DP-SVRG, is used or not.
In some scenarios that $p \gg d$ (such as neural network), this gap can be ignored.
Meanwhile, the excess empirical risk bound of our method is much better than the input perturbation method proposed in \cite{62}, approximately by a factor of $\frac{1}{n}$, which is a huge gap.
Considering that the ($O(\sqrt{n\epsilon})$,$\delta$)-LDP guaranteed by the input perturbation method proposed in \cite{62} is unsatisfactory (actually, this privacy is really weak because $n$ is always up to hundreds or thousands), the sacrifice on LDP for the improvement on performance in our method is acceptable.

In this paper, we add noise to data, leading the perturbation on the gradient and achieves ($\epsilon$,$\delta$)-differential privacy on the model parameter, building a bridge between local and central differential privacy.
By detailed analysis, it can be observed that the theoretical results of our method are similar to (or even better than) previous central perturbation methods.
Experimental results also show that the performance of our proposed method is similar to the gradient perturbation method proposed in \cite{20} and the output perturbation method proposed in \cite{22}.
Our method preserves the privacy of the gradient, ($\epsilon$,$\delta$)-differential privacy on final model parameters along with some kind of original data privacy, without decreases on theoretical or practical results, which is an attractive result.

\section{PRELIMINARIES}
In this section, first, we introduce some basic definitions, including the comparison between central and local differential privacy.
Then, we list traditional perturbation methods of central differentially private ERM in detail: output perturbation, objective perturbation and gradient perturbation.

\subsection{Notations and Basic Definitions}
Given a $d$-dimensional vector $\mathbf{x}$=$[x_1,x_2,...,x_d]^\top$, denotes its $\ell_2$-norm by $\left\| \mathbf{x}  \right\|$=$(\sum_{i=1}^{d} \vert x_i \vert^2)^{\frac{1}{2}}$. Two databases $D,D' \in \mathcal{D}^n$ differing by one element are denoted by $D \sim D'$, called \textit{adjacent databases}.

\begin{Def}[Central Differential Privacy \cite{27}]
A randomized function $\mathcal{A} : \mathcal{D}^n \rightarrow \mathbb{R}^p$ is ($\epsilon$,$\delta$)-differential privacy if
\begin{equation}\label{CDP}
\mathbb{P}[\mathcal{A}(D) \in S] \leq e^\epsilon \mathbb{P}[\mathcal{A}(D') \in S] + \delta,
\end{equation}
where $S \in$ range($\mathcal{A}$) and $p$ is the number of parameters.
\end{Def}

\begin{Def}[Local Differential Privacy \cite{46}]
An algorithm $\mathcal{Q}$ is ($\epsilon$,$\delta$)-local differential privacy if for all $x,x' \in D$, and for all events $E$ in the output space of $\mathcal{Q}$, we have:
\begin{equation}\label{LDP}
\mathbb{P}[\mathcal{Q}(x) \in E] \leq e^\epsilon \mathbb{P}[\mathcal{Q}(x') \in E] + \delta.
\end{equation}
\end{Def}

According to the definitions of central and local differential privacy, in Definition 1, datasets $D$ and $D'$ are input to the randomized function $\mathcal{A}$, the privacy of the machine learning model is focused, guaranteeing information cannot be inferred by observing the \textbf{machine learning model}.
In Definition 2, records $x$ and $x'$ are input to the algorithm $\mathcal{Q}$, data is paid more attention, guaranteeing information cannot be inferred by observing the `\textbf{noisy data}'.
In the local model, `untrusted server' is seemed as the malicious adversary.

\subsection{Traditional Perturbation Methods}
Our method focuses more on the privacy of the machine learning model, similar to the central setting.
So, in this part, we introduce three traditional central perturbation methods.

In general, the objective function of ERM without privacy preserving is defined as:
\begin{equation}\label{ERMObj}
L(\theta)=\frac{1}{n}\sum_{i=1}^{n} \ell(\theta,x_i,y_i),
\end{equation}
where $(x_i,y_i)$ denotes data instance, $\ell(\cdot)$ is the loss function.

In the case of binary classification, the data space $\mathcal{X}=\mathbb{R}^d$ and the label set $\mathcal{Y}=\{-1,+1\}$, and we assume throughout that $\mathcal{X}$ is the unit ball so that $\|x_i\| \leq 1$.

\begin{OP}
\rm{
In output perturbation, noise is directly added to the model (in the paper, we denote model by parameters):
\begin{equation}\label{output}
\theta_{priv}=\arg\min \left[L(\theta)\right]+z,
\end{equation}
where $z$ is the noise guaranteeing differential privacy.
}
\end{OP}
Output Perturbation method is commonly used because it is simple to implement, only adding noise to the final model.

\begin{ObP}
\rm{
In the method of objective perturbation, noise is added to the objective function:
\begin{equation}\label{objective}
L_{priv}(\theta)=L(\theta)+\frac{1}{n}z^{T}\theta.
\end{equation}
}

The perturbed objective function $L_{priv}(\theta)$ is directly optimized:
\begin{equation}\label{objopt}
\theta_{priv}=\arg\min \left[L_{priv}(\theta)\right].
\end{equation}
Note that in (\ref{objective}), there may be some other terms on the right side of the equality, for example $\frac{\Delta}{2}\|\theta\|^2$ in \cite{23}.
We only list the most important term $\frac{1}{n}z^{T}\theta$ to guarantee differential privacy here.
\end{ObP}
This method is rarely used in recent years because it is always a trouble to optimize the perturbed objective function and the performance is unsatisfactory.

\begin{GP}
\rm{
In the gradient perturbation method, noise is added to the gradient when training, which leads the gradient descent process at round $t$ to:
\begin{equation}\label{gradient}
\theta_{t+1}=\theta_{t}-\alpha(\nabla L(\theta_{t})+z),
\end{equation}
where $\alpha$ is the learning rate.

After $T$ iterations in total, the final model $\theta_{priv}=\theta_{T}$.
}
\end{GP}
Because most machine learning algorithms are based on gradient descent method, gradient perturbation is feasible and popular.

\section{DIFFERENTIALLY PRIVATE ERM WITH INPUT PERTURBATION}
In this section, first, we analyze the weaknesses of traditional central perturbation methods and local models introduced in Section 3, then we propose our method \textit{input perturbation} in detail.

When training models, original data is always sent to the `data center' in advance, which is shown in Figure 1.
By observing three traditional perturbation methods of central DP-ERM, original data is not protected, which means the `data center' is assumed trusted.

However, `data center' is not easy to `trust' because the adversaries always desire to `take away' the original data and the `data center' may be monitored with high probability.
As a result, the security of original data instances is of the same importance as (or even more important than) the model parameters.
LDP is a superior way to solve the problem of `untrusted data center', guaranteeing differential privacy over the communications (data exchanging) between individuals and the `data center'.
However, as shown in Table 1, the noise added to data is large, and it is inevitable that the performance is worse than central models.

To solve the problems mentioned above, we propose a new input perturbation method, adding noise to data instances and training the machine learning model by the `perturbed data instances', which leads the objective function to:
\begin{equation}\label{inputobj}
\hat{L}(\theta)=\frac{1}{n}\sum_{i=1}^{n} \ell(\theta,x_i+z,y_i).
\end{equation}

In order to distinguish with the objective function without privacy consideration $L(\theta)$ in (\ref{ERMObj}), we denote the objective function of input perturbation by $\hat{L}(\theta)$.
In (\ref{inputobj}), `noise adding' has been done in advance and the formulation $x_i+z$ is for distinguishing the perturbed data and original data.

Our method focuses on achieving ($\epsilon$,$\delta$)-differential privacy on the machine learning model with some kind of privacy on original data.
As a result, even if the `data center' is not trusted or monitored, the data `taken away' by malicious adversaries is with random noise, which preserves the `true original data' of individuals from some kinds of attacks.

Although in our method, noise is added to the original data, we focus more on the ($\epsilon$,$\delta$)-differential privacy of the final model, which is different from the local model: protections between individuals and the `server' are paid more attentions, and the privacy of model parameters is not discussed.
Comparing with LDP and input perturbation method in \cite{62}, based on the aim to guarantee the quality of the machine learning model, we sacrifice some of the privacy on individuals for the performance.
In fact, the sacrifice compared with \cite{62} is not much.
In other words, focusing on keeping good performance, we attempt to preserve the privacy on original data as much as possible.
It can be observed that in LDP and previous input perturbation method, the noise added to data is much more than ours.
As a result, the privacy preserving on individuals of our method is weaker than in LDP and previous input perturbation method, but still stronger than central methods.

Our method is detailed in Algorithm 1.

\begin{algorithm}
    \caption{Differentially Private ERM with Input Perturbation Method}
    \begin{algorithmic}[1]
        \Require Dataset $D$, iteration rounds $T$, learning rate $\alpha$
        \Function {InputPerturbation}{$D,T,\alpha$}
        \State For all data instances $(x_i,y_i)$ in $D$, add noise $z$ to it:
        \State $(x_i,y_i) \leftarrow (x_i+z,y_i)$.
        \State New data $(x_i+z,y_i)$ is denoted as `perturbed data'.
        \State Train model by perturbed data, the objective function is the same as (\ref{inputobj}), which leads the following process.
        \State \textbf{for} $t=0$ to $T-1$ \textbf{do}
        \State \quad $\theta_{t+1} \leftarrow \theta_{t}-\alpha\frac{1}{n}\sum_{i=1}^{n}\nabla\hat{L}(\theta_{t})$.
        \State \textbf{end for}
        \State return $\theta_{T}$.
        \EndFunction
    \end{algorithmic}
\end{algorithm}

In Algorithm 1, the random noise $z \sim \mathbb{R}^d$ and each element $z_i \sim \mathcal{N}(0, \sigma^2)$, sampled independently.
By line 7 in Algorithm 1, it can be seen that the noise added to the original data affects the gradient.
The theoretical analysis of our method in Section 5 is based on this observation.

Besides, by observing that our method adds noise to original data instances, leading perturbation on the gradient and eventually causing perturbation on the model parameters, a bridge is built between local and central differential privacy: input perturbation ERM protects the original data, the gradient and the final model simultaneously, giving a higher level privacy compared with traditional central perturbation methods without decreases on the theoretical or practical results.
Meanwhile, we achieve better performance compared with LDP and previous input perturbation method, by sacrificing some amount of privacy on individuals.

\section{THEORETICAL ANALYSIS OF INPUT PERTURBATION ERM}
In this section, first, we give privacy guarantees of our proposed method: input perturbation ERM.
Then, we analyze the excess empirical risk bound of our method.
Finally, we extend our method to a more general case, in which the loss function is not restricted strongly convex but satisfies the Polyak-Lojasiewicz condition, which is more general than the property `strongly convex'.

\begin{figure*}[htbp]
\centering
\subfigure[KDDCup99 (LR)]{\includegraphics[width=0.3\textwidth]{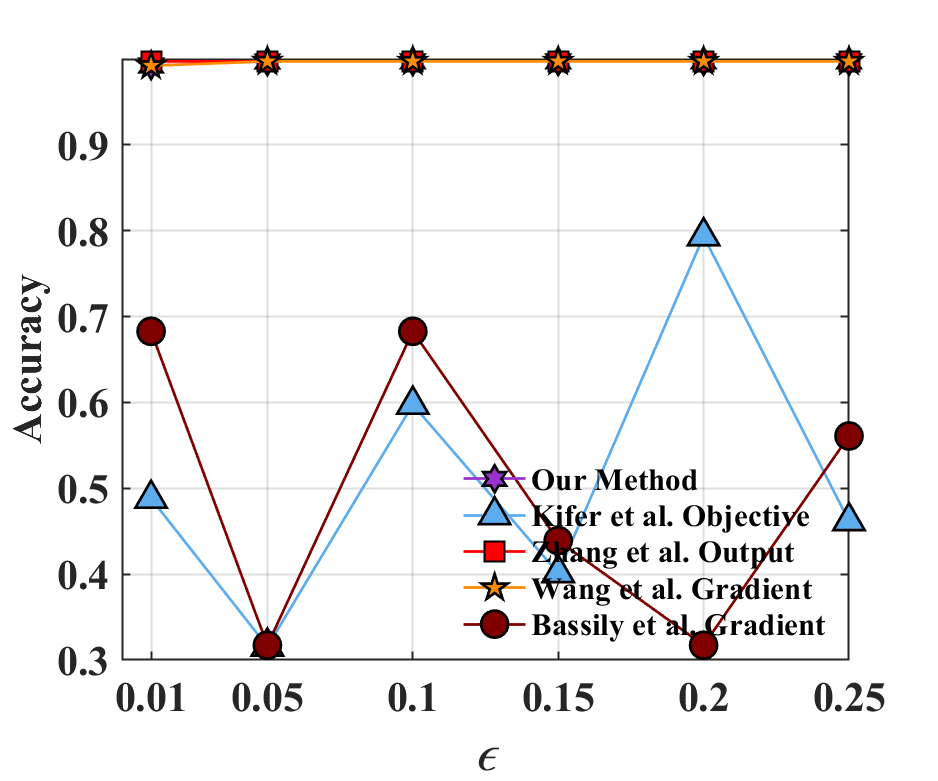}}
\subfigure[Adult (LR)]{\includegraphics[width=0.3\textwidth]{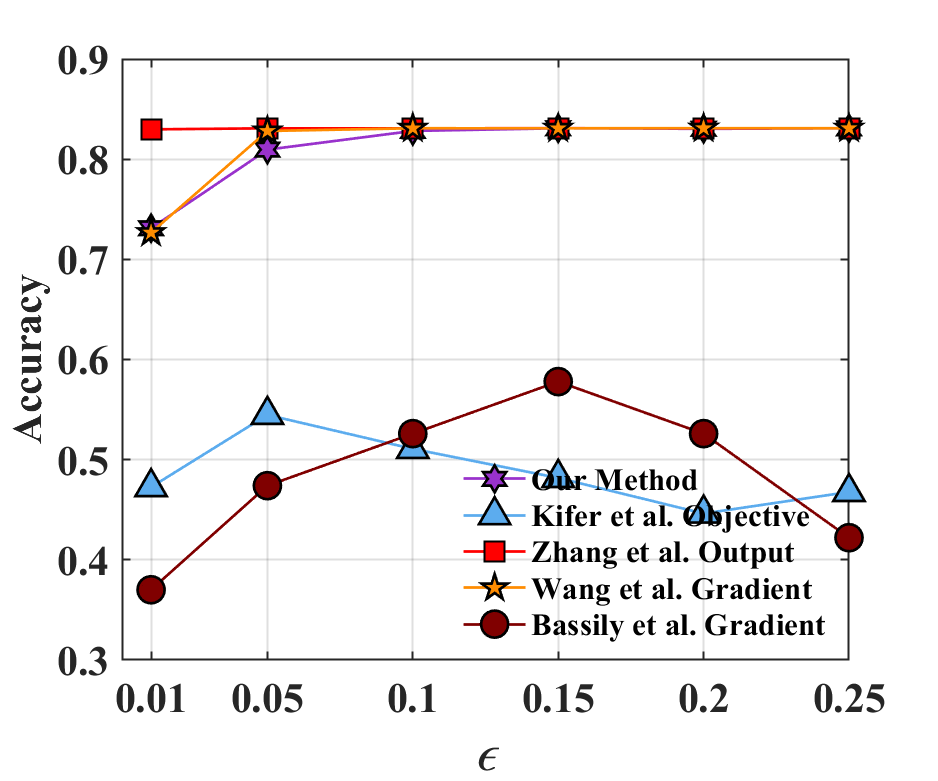}}
\subfigure[Bank (LR)]{\includegraphics[width=0.3\textwidth]{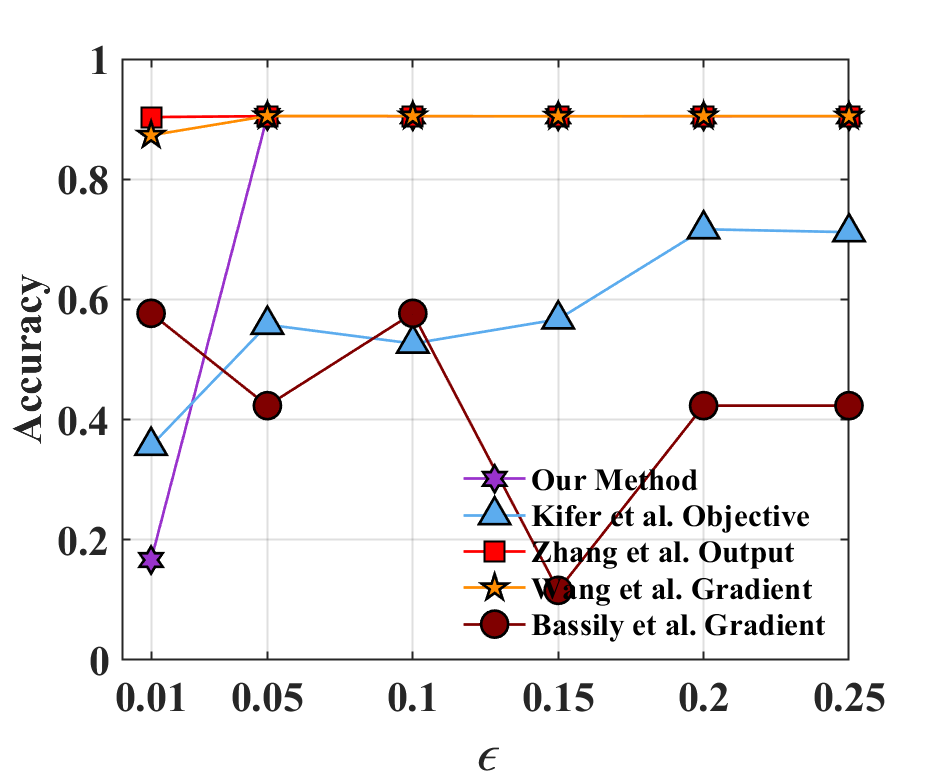}}
\subfigure[KDDCup99 (MLP)]{\includegraphics[width=0.3\textwidth]{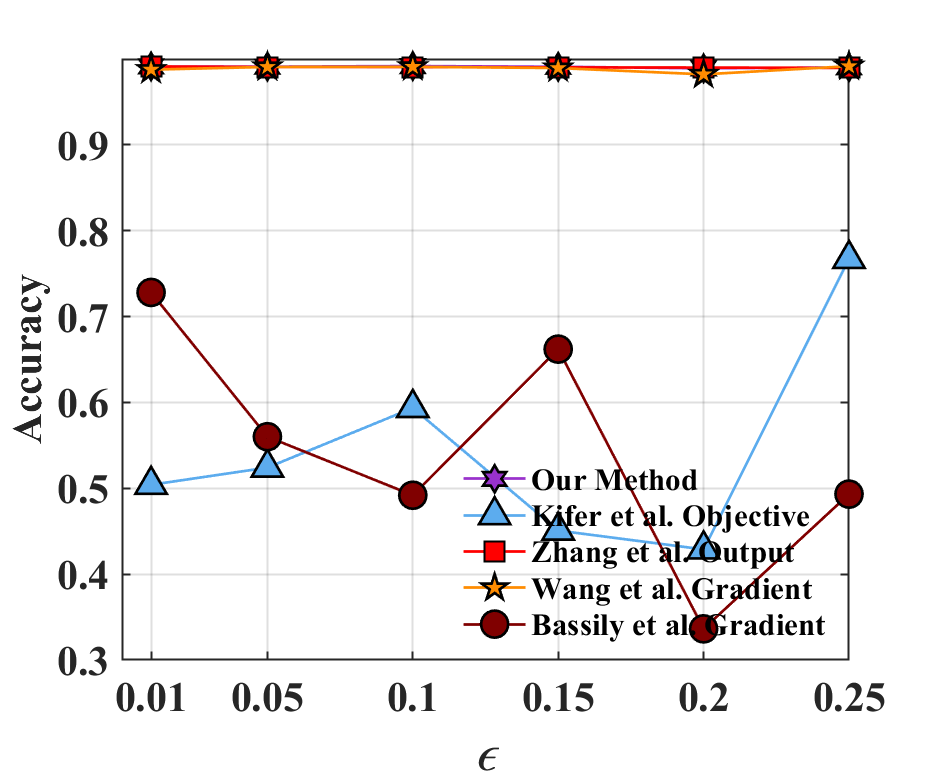}}
\subfigure[Adult (MLP)]{\includegraphics[width=0.3\textwidth]{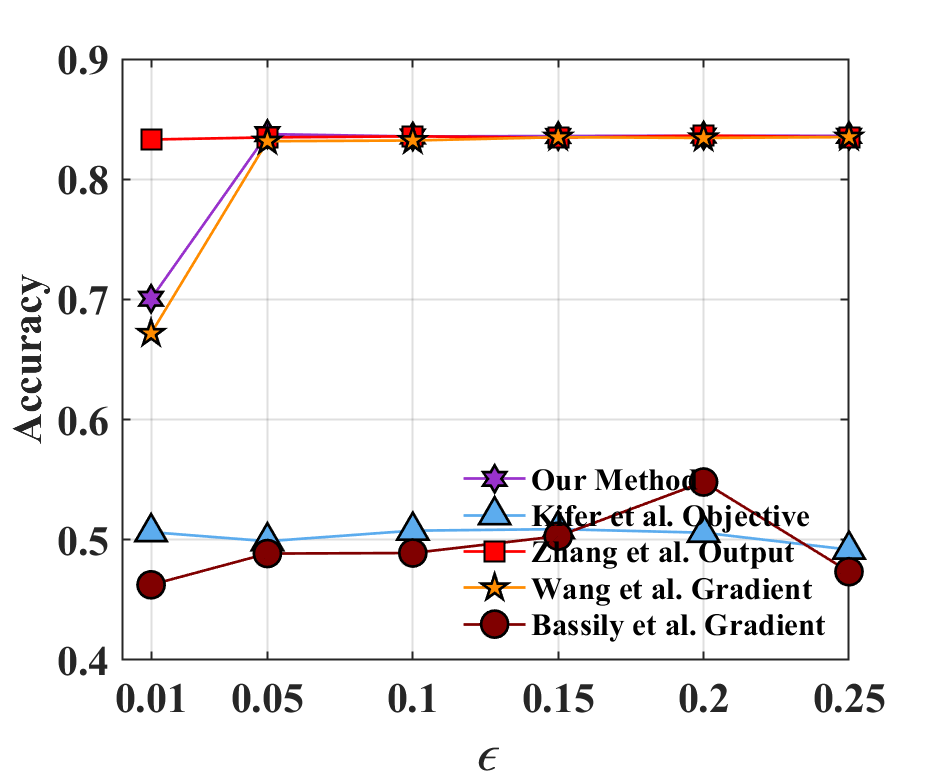}}
\subfigure[Bank (MLP)]{\includegraphics[width=0.3\textwidth]{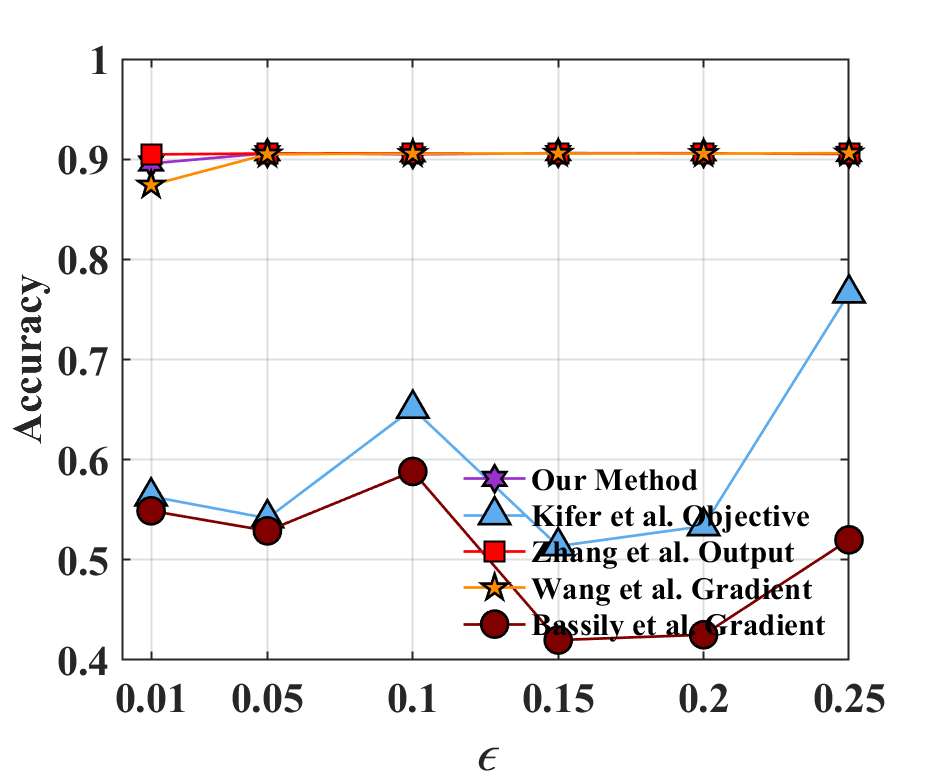}}
\caption{Accuracy over $\epsilon$ on different datasets.}
\end{figure*}

\begin{figure*}[htbp]
\centering
\subfigure[KDDCup99 (LR)]{\includegraphics[width=0.3\textwidth]{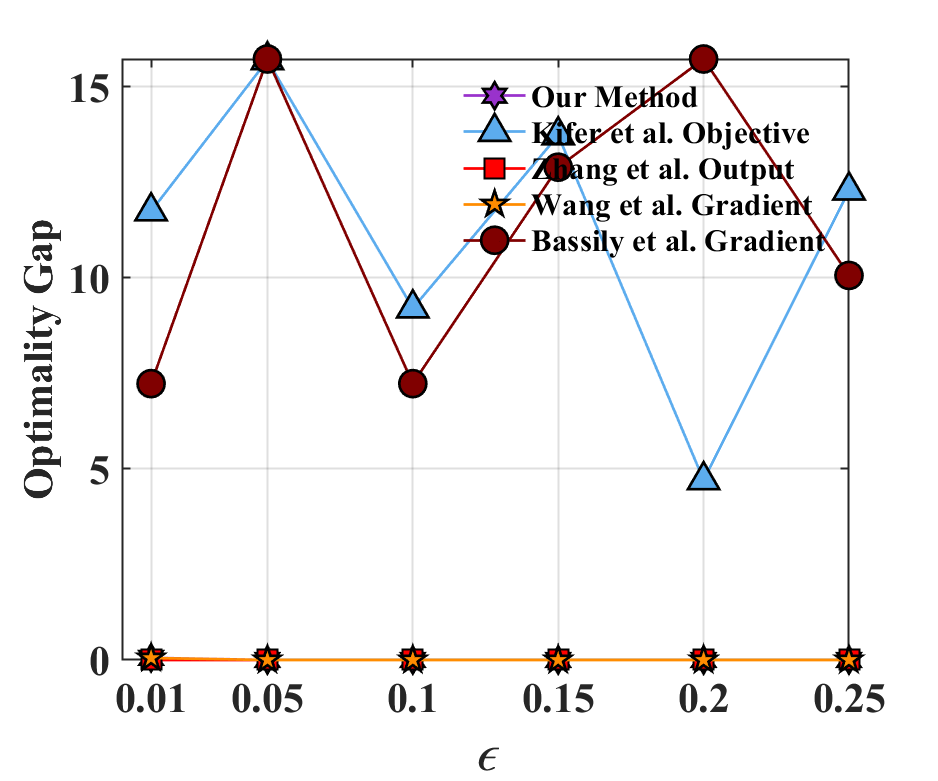}}
\subfigure[Adult (LR)]{\includegraphics[width=0.3\textwidth]{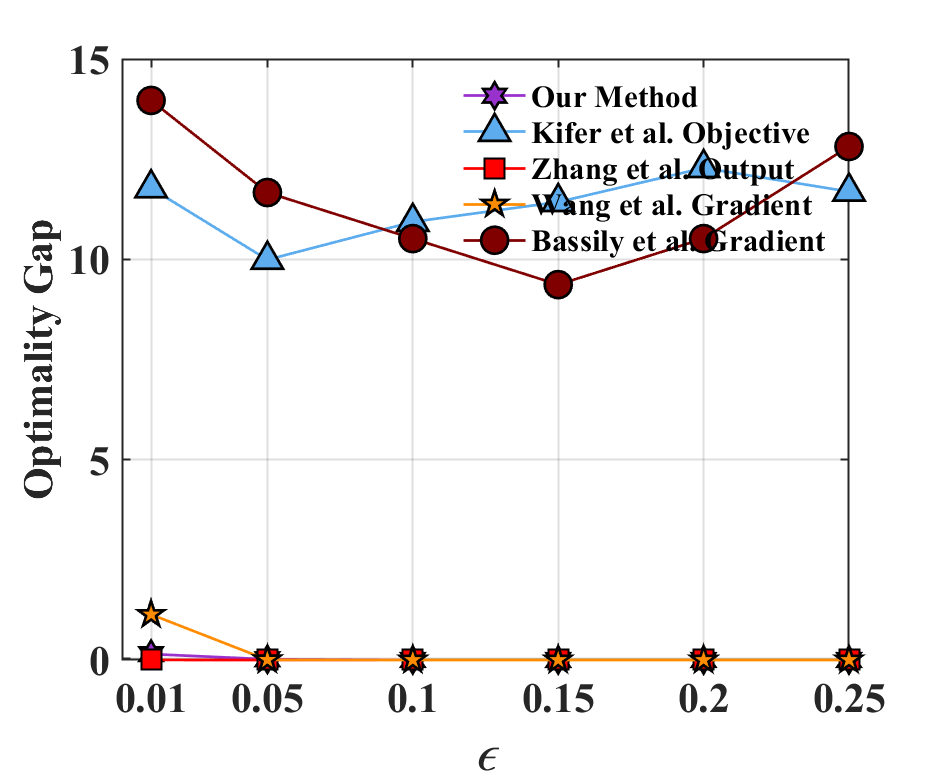}}
\subfigure[Bank (LR)]{\includegraphics[width=0.3\textwidth]{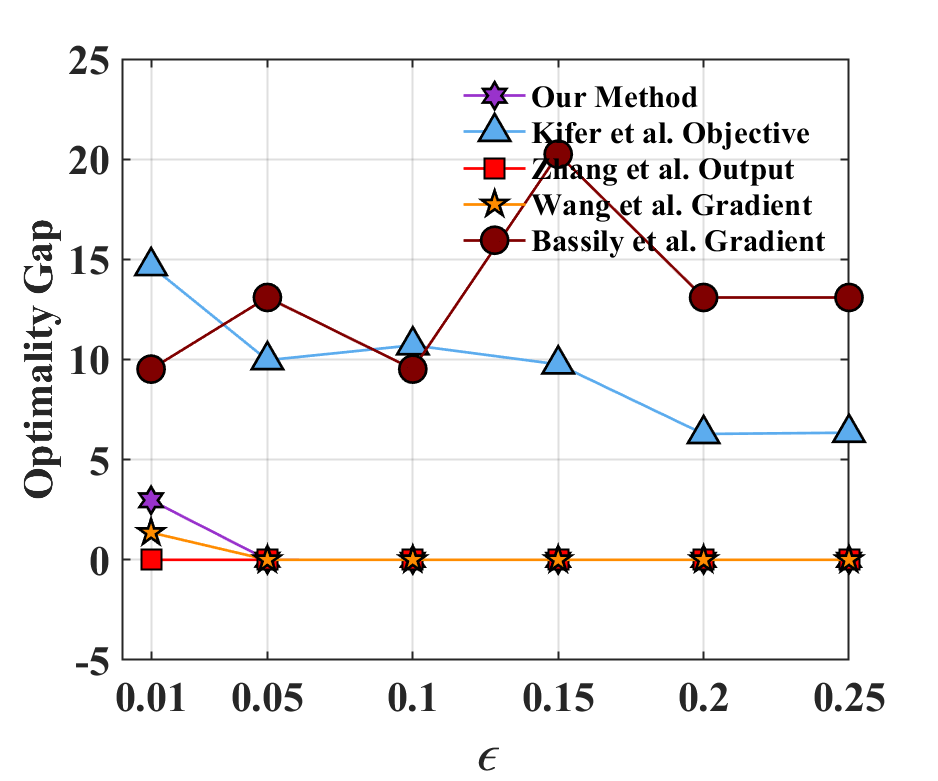}}
\subfigure[KDDCup99 (MLP)]{\includegraphics[width=0.3\textwidth]{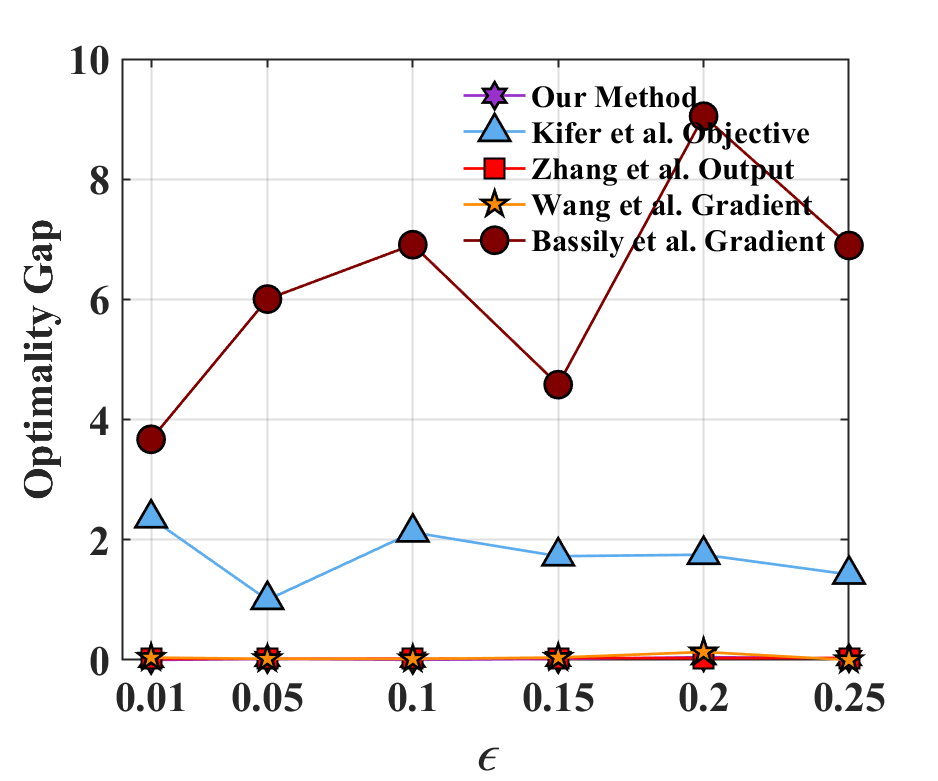}}
\subfigure[Adult (MLP)]{\includegraphics[width=0.3\textwidth]{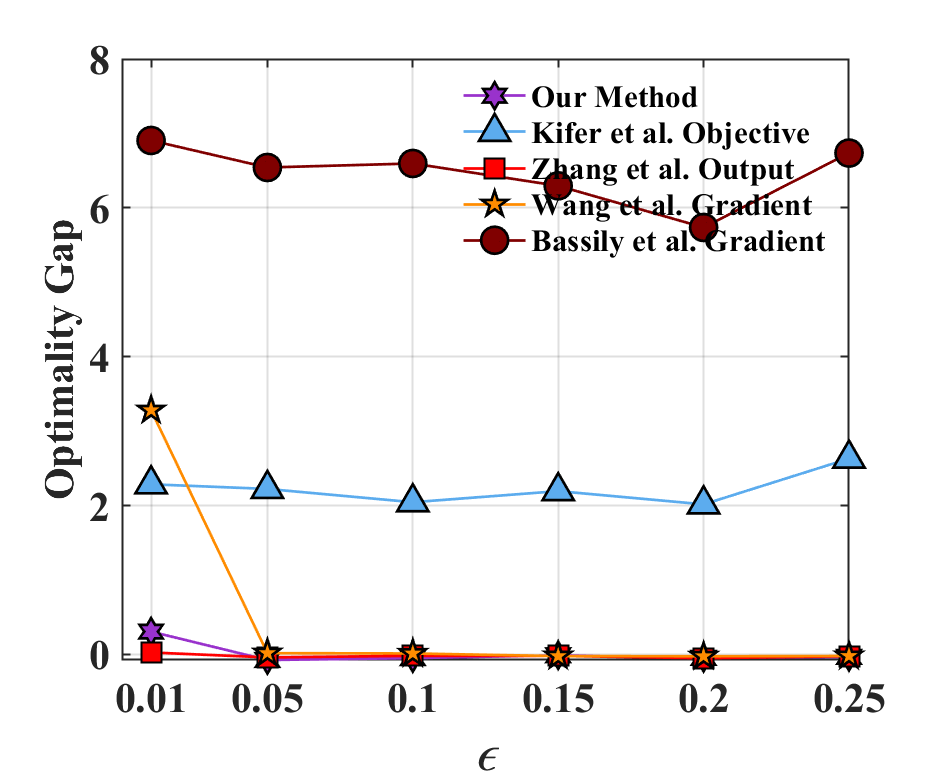}}
\subfigure[Bank (MLP)]{\includegraphics[width=0.3\textwidth]{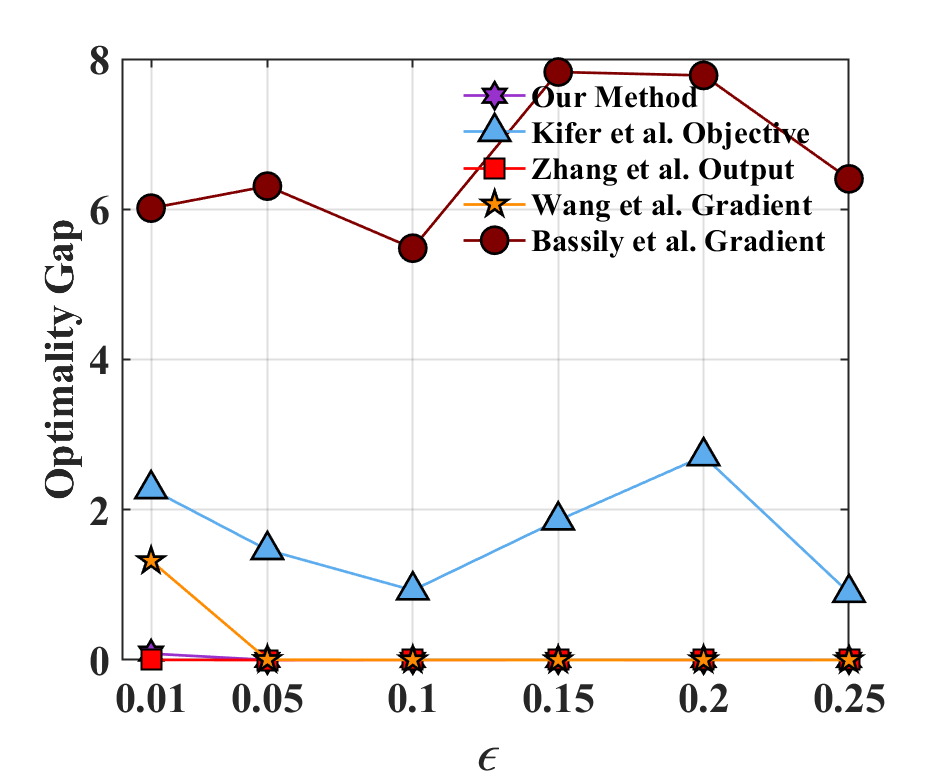}}
\caption{Optimality gap over $\epsilon$ on different datasets.}
\end{figure*}

\begin{figure*}[htbp]
\centering
\subfigure[Breast Cancer (LR)]{\includegraphics[width=0.3\textwidth]{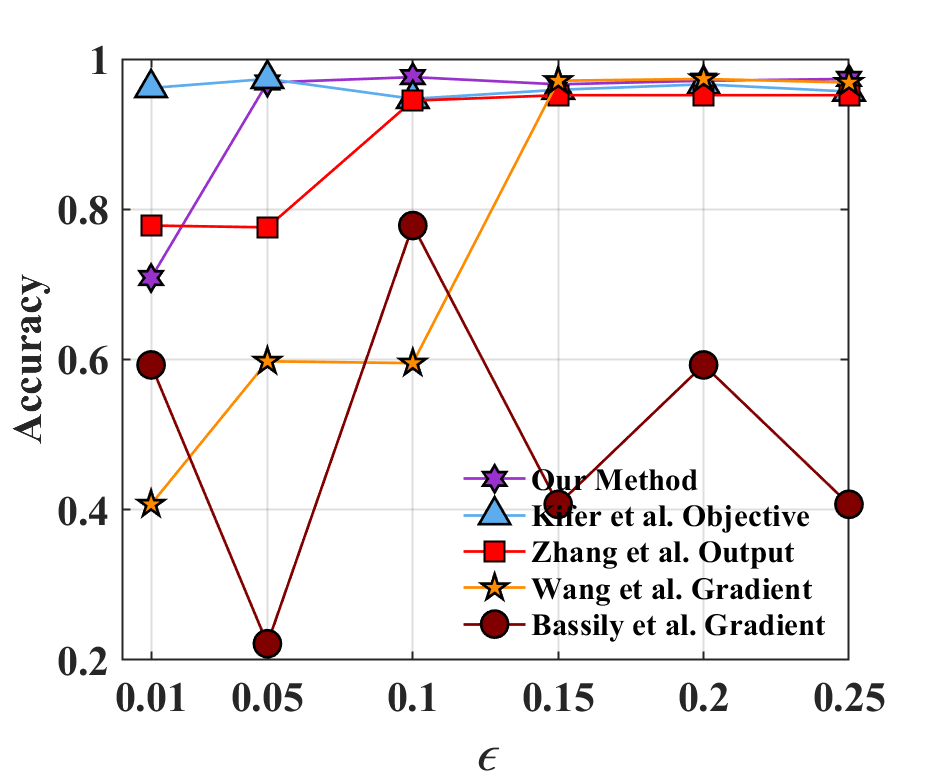}}
\subfigure[Credit Card Fraud (LR)]{\includegraphics[width=0.3\textwidth]{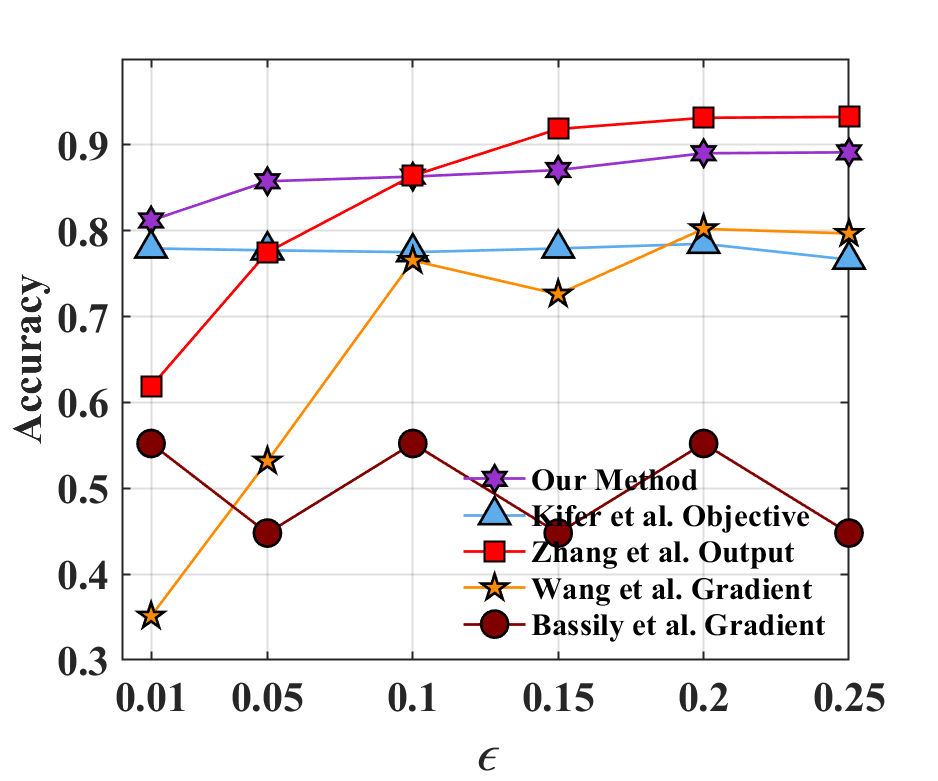}}
\subfigure[Iris (LR)]{\includegraphics[width=0.3\textwidth]{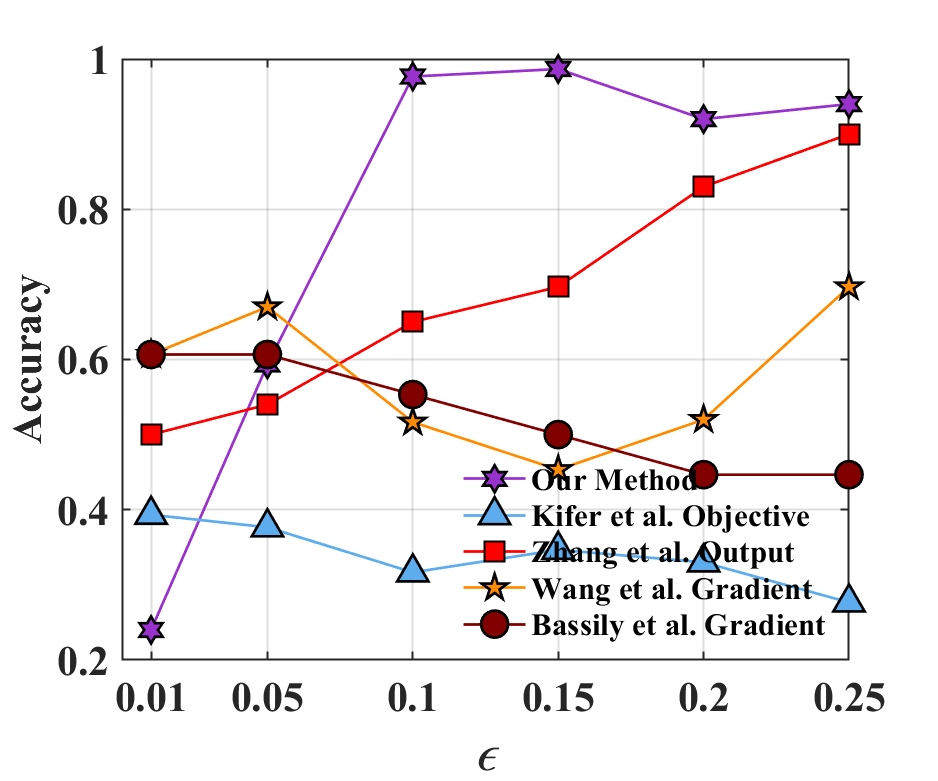}}
\subfigure[Breast Cancer (LR)]{\includegraphics[width=0.3\textwidth]{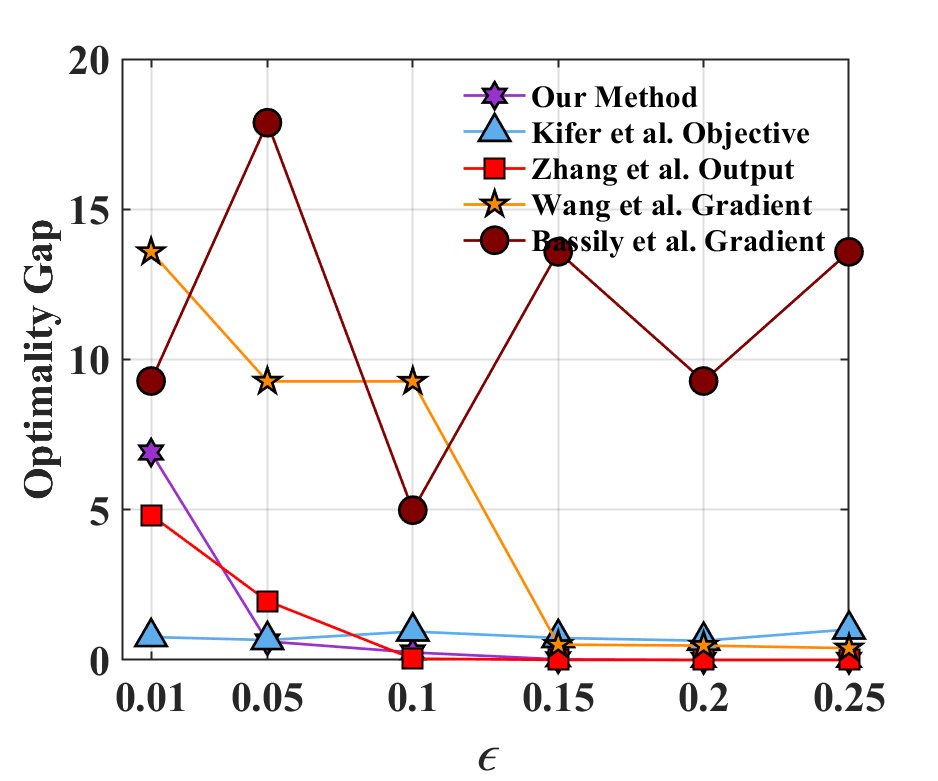}}
\subfigure[Credit Card Fraud (LR)]{\includegraphics[width=0.3\textwidth]{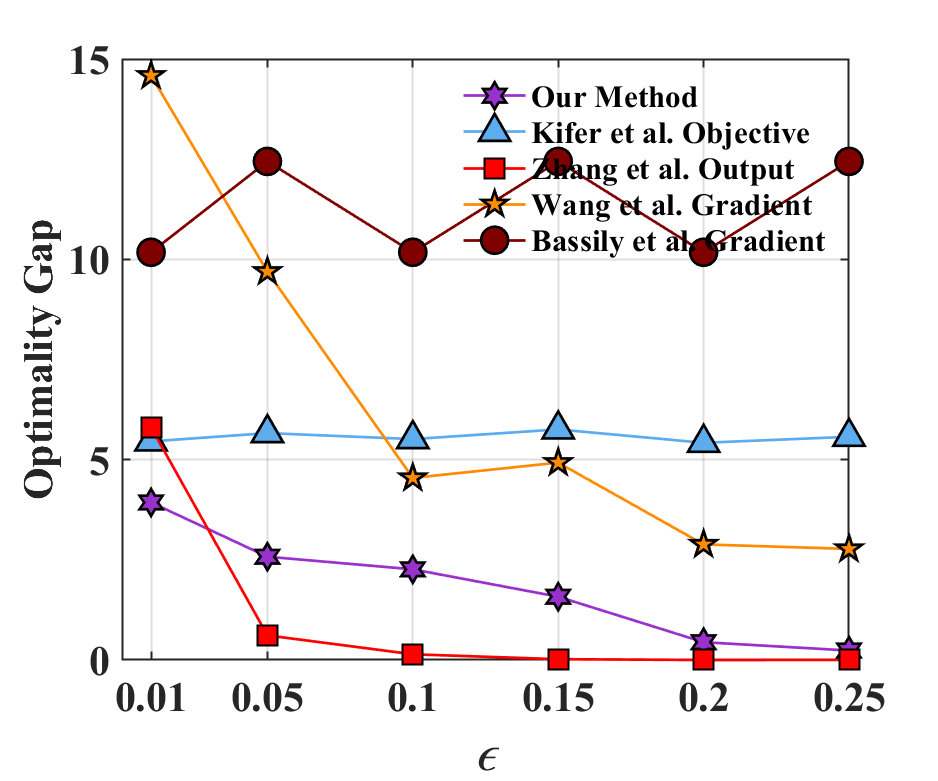}}
\subfigure[Iris (LR)]{\includegraphics[width=0.3\textwidth]{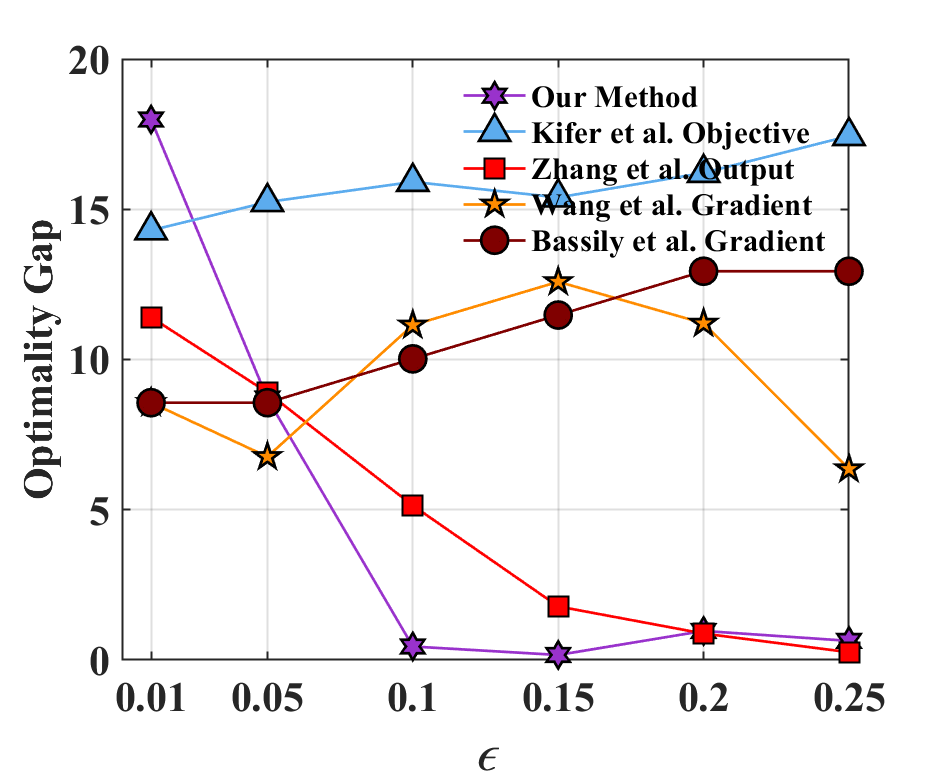}}
\caption{Accuracy and optimality gap over $\epsilon$ on different datasets.}
\end{figure*}

\subsection{Differential Privacy}
In this part, we analyze the ($\epsilon$,$\delta$)-differential privacy of our proposed method: input perturbation in Algorithm 1.

In this paper, we analyze $(\epsilon,\delta)$-differential privacy by Gaussian mechanism proposed in \cite{8} and moments accountant proposed in \cite{15}.
Moreover, we assume $\ell(\theta,x,y)$ is $\ell(y\theta^{T}x)$ like in \cite{18}.

\begin{The}
In Algorithm 1, for $\epsilon,\delta>0$, if $\ell(\theta,x,y)$ is $G$-Lipschitz and $\Delta$-strongly convex over $\theta$ and
\begin{equation}\label{the1}
\sigma^2=c\frac{G^2T\log(1/\delta)}{n(n-1)\sqrt{\Delta}\epsilon^2},
\end{equation}
it is ($\epsilon$,$\delta$)-differential privacy for some constant $c$.
\end{The}
The proof is detailed in the Appendix.

It can be observed that by using our method, the noise added to data instances is almost the same as the gradient perturbation method proposed in \cite{20}.
The difference is by a factor of $\frac{(n-1)\sqrt{\Delta}}{n}$, which can be seen as a constant.
When comparing with the traditional gradient perturbation method proposed in \cite{21}, our noise bound is much better than it by a factor up to $\frac{n^4\log(n)}{T}$.
Meanwhile, the noise bound of our method is far better than LDP methods, considering that LDP preserves stronger privacy between individuals and the `server', and our method pays more attentions on the privacy of the final machine learning model, this result is conceivable.

The similarity between our method and the gradient perturbation method is the same as our observation: perturbation on original data causes the perturbation on gradients, which builds a bridge between local and central differential privacy.
As a result, our proposed input perturbation method achieves ($\epsilon$,$\delta$)-DP on the final model through this `bridge'.
Hence, our method preserves the privacy of the original data instances, the gradient and the model parameters simultaneously, providing a higher level protection on privacy in a more reliable way in the field of central DP-ERM.

\subsection{Excess Empirical Risk Bound}
In this part, we analyze the utility of our proposed method and give the excess empirical risk bound, denoted by the expectation of $\hat{L}(\theta_{T})-L^{*}$, where $L^{*}$ is the value of the objective function over the optimal model without privacy consideration.
Formally, $L^{*}=\min_{\theta\in\mathbb{R}^{p}}L(\theta)$, where $L(\theta)$ is the same as in (\ref{ERMObj}).

\begin{The}
Suppose that $\ell(\theta,x,y)$ is $G$-Lipschitz, $\nabla\ell(\theta,x,y)$ is $L$-Lipschitz\footnote[5]{$L$-Lipschitz on $\nabla\ell(\cdot)$ means $L$-smooth on $\ell(\cdot)$.} and the $\ell_2$-norm of the model parameter has an upper bound $D$ (i.e. $\|\theta\| \leq D$ for all $\theta$), with $\sigma$ is the same as in (\ref{the1}), we have:
\begin{equation}\label{the2}
\mathbb{E}\left[\hat{L}(\theta_{T})-L^{*}\right] \leq O\left(\frac{\alpha(2LD+G)G^3d\log^2(n)\log(1/\delta)}{n(n-1)\sqrt{\Delta}\epsilon^2}\right),
\end{equation}
where $T=\widetilde{O}\left(\log(\frac{n(n-1)\sqrt{\Delta}\epsilon^2}{\alpha(2LD+G)G^3d\log(1/\delta)})\right)$, $\alpha$ represents the learning rate and each data instance $x_i \in \mathcal{X} \in \mathbb{R}^{d}$ has $d$-dimensional features.
\end{The}

The proof is shown in the Appendix.

\begin{Rem}
Considering that the smoothness of the objective function after input perturbation $\hat{L}(\theta)$ is not easy to achieve because of the existence of the random variable $z$, we assume $L(\theta)$ (without random variables) is $L$-smooth, which is easier to hold, making the utility and the excess empirical risk bound of our method feasible.
\end{Rem}

It can be observed that the excess empirical risk bound of our method is better than the traditional gradient perturbation method proposed in \cite{21} by a factor of $\frac{\alpha(2LD+G)Gd\sqrt{\Delta}}{np}$.
Considering that the variables $L,D,G,\alpha,\Delta$ can be seemed as constants, our method in much better than which proposed in \cite{21} by a factor of $\frac{d}{np}$.
When comparing with gradient perturbation methods proposed in \cite{20}, the gap on empirical risk bound is by a factor of $\frac{d\log(n)}{p}$.
In some cases that $p \gg d$, which is common in the field such as deep learning, the gap between our method and the gradient perturbation methods proposed in \cite{20} is relatively small and can be ignored.
When it comes to the comparison between our method and LDP methods, the excess empirical risk bound of our method is much better, with weaker privacy on individuals.

\subsection{More general condition}
In this part, we extend our method to a more general condition that the loss function $\ell(\theta,x,y)$ is not restricted $\Delta$-strongly convex, but satisfies the Polyak-Lojasiewicz condition.
\begin{Def}
Given a function $\ell(\cdot)$, if there exists $\mu > 0$ and for all $\theta$, we have:
\begin{equation}\label{PL}
\|\nabla\ell(\theta)\|^2 \geq 2\mu(\ell(\theta)-\ell^{*}),
\end{equation}
then $\ell(\cdot)$ satisfies the Polyak-Lojasiewicz condition.
\end{Def}
The Polyak-Lojasiewicz condition is much more general than \textit{strongly convex}.
It was shown in \cite{30} that when function $\ell$ is differential and $L$-smooth under $\ell_2$-norm, we have:

Strong Convex $\Rightarrow$ Essential Strong Convexity $\Rightarrow$ Weak Strongly Convexity $\Rightarrow$ Restricted Secant Inequality $\Rightarrow$ Polyak-Lojasiewicz Inequality $\Leftrightarrow$ Error Bound

\begin{The}
In Algorithm 1, for $\epsilon,\delta>0$, if the loss function $\ell(\theta,x,y)$ is $G$-Lipschitz and satisfies Polyak-Lojasiewicz condition over $\theta$ and
\begin{equation}\label{the3}
\sigma^2=c\frac{G^2T\log(1/\delta)}{n(n-1)\epsilon^2},
\end{equation}
it is ($\epsilon$,$\delta$)-differential privacy for some constant $c$.
\end{The}

Detailed proof is shown in the Appendix.

\begin{The}
Suppose that $\ell(\theta,x,y)$ is $G$-Lipschitz, $\nabla\ell(\theta,x,y)$ is $L$-Lipschitz, $L(\theta)$ is $L$-smooth over $\theta$ and the $\ell_2$-norm of the model parameter has an upper bound $D$ (i.e. $\|\theta\| \leq D$), with $\sigma$ is the same as in (52), we have:
\begin{equation}\label{the4}
\mathbb{E}\left[\hat{L}(\theta_{T})-L^{*}\right] \leq O\left(\frac{\alpha(2LD+G)G^3d\log^2(n)\log(1/\delta)}{n(n-1)\epsilon^2}\right),
\end{equation}
where $T=\widetilde{O}\left(\log(\frac{n(n-1)\epsilon^2}{\alpha(2LD+G)G^3d\log(1/\delta)})\right)$, $\alpha$ is the learning rate and each data instance $x_i$ has $d$-dimensional features.
\end{The}

The proof of Theorem 4 is almost the same as Theorem 2, with replacement of $\sigma$.

By Theorem 3 and Theorem 4, it can be observed that in a more general case: the loss function is not restricted strongly convex but satisfies the Polyak-Lojasiewicz condition, our noise bound and the excess empirical risk bound are almost the same as previous work on central models.

\section{EXPERIMENTS}
The experiments are performed on the classification task.
Considering that our method focuses on the privacy of the final model, the experiments are applied on central methods: the objective perturbation method proposed in \cite{23}, the output perturbation method proposed in \cite{22} and the gradient perturbation methods proposed in \cite{21} and \cite{20} (without DP-SVRG).
The performance is represented by accuracy and the optimality gap, the latter is defined as $L(\theta_{priv})-L^{*}$.
Accuracy represents the performance on test data and optimal gap denotes excess empirical risk on training data.

According to the sizes of datasets, we use logistic regression model (LR) and deep learning model on the datasets KDDCup99 \cite{31}, Adult \cite{32}, Bank \cite{33}, where the total number of data instances are 70000, 45222 and 41188, the sizes are large than 10000.
On datasets Breast Cancer \cite{34}, Credit Card Fraud \cite{35}, Iris \cite{32}, only logistic regression model is applied because the sizes are less than 1000, where the total number of data instances are 699, 984 and 150, respectively.
In the experiments, deep learning model is denoted by Multi-layer Perceptron (MLP) with one hidden layer whose size is the same as the input layer.
The training set and the testing set are chosen randomly.

In all experiments, $T$ and $\alpha$ are chosen by cross-validation.
We evaluate the influence over differential privacy budget $\epsilon$, which is set from 0.01 to 0.25.
Meanwhile, $\delta$ is set according to the size of datasets and can be seemed as a constant.
Note that in logistic regression model, $d=p$ and in deep learning model, $d<p$.

Figure 2 shows that the accuracy of our proposed method is better than the gradient perturbation method proposed in \cite{21} and the objective perturbation method proposed in \cite{23}.
And our method is almost the same as the gradient perturbation method proposed in \cite{20} and the output perturbation method proposed in \cite{22} on accuracy, no matter on the LR model or on the MLP model.
However, because the variance of the Gaussian noise added to the gradient in the method \cite{21} is large: $O\left(\frac{G^2n^2\log(n/\delta)\log(1/\delta)}{\epsilon^2}\right)$, the accuracy of this method over $\epsilon$ fluctuates sharply in Figure 2.

It can be observed that in Figure 3, the optimality gap of our method is almost the same as the output perturbation method proposed in \cite{20} and is better than other methods mentioned above over most datasets, which is similar to the theoretical analysis.
Moreover, it can be observed that the optimality gap of our method on some datasets are close to 0, which means that our method achieves almost the same performance as the ERM model without privacy consideration in some scenarios, on both LR model and MLP model.
In addition, like the accuracy in Figure 2, the optimality gap of the gradient perturbation method proposed in \cite{21} fluctuates sharply because of its noise bound.

Figure 4 shows accuracy and optimality gap on small datasets (the sizes are less than 1000), in which only logistic regression model is applied.
The results are similar to which in Figure 2 and Figure 3, which means that our method is effective in most cases.

By observing the experimental results, we find that although there are slight differences in experimental results on different datasets, the performance of the gradient perturbation method proposed in \cite{21} and the objective perturbation method proposed in \cite{23} is much weaker than our method, the former is because of its loose noise bound and the latter is because of the perturbation method itself.
Our proposed method: input perturbation, is almost the same as (on some datasets, even better than) the output perturbation method in \cite{20} and the traditional gradient perturbation method without DP-SVRG in \cite{22} on both accuracy and optimality gap, which is similar to our theoretical analysis in Section 4.
The experimental results on the deep learning model (MLP), are similar to the traditional machine learning (logistic regression) model.
Considering that our method preserves the privacy of the original data, the gradient and the final model simultaneously, providing more privacy without decreases on the performance compared with previous central methods, it is an attractive result.

\section{CONCLUSIONS}
In this paper, we study the input perturbation method in DP-ERM, adding Gaussian noise to original data instances and training the machine learning model by the `perturbed data'.
By observing that input perturbation leads the perturbation on the gradient and finally the perturbation on the final model, we build a bridge between local and central differential privacy, achieving ($\epsilon$,$\delta$)-differential privacy on the final machine learning model, along with some kind of privacy on individuals.
Through the `bridge', we preserve the original data, the gradient and the final machine learning model simultaneously.
Meanwhile, we extend our method to a more general condition, in which the loss function is not considered strongly convex but satisfies the Polyak-Lojasiewicz condition.
Theoretical analysis and experiments (applied on both traditional machine learning model: logistic regression, and deep learning model: MLP) on real datasets show that our method achieves almost the same (or even better) performance compared with some of the best previous methods.
Additionally, higher level of privacy is achieved, comparing with previous central methods.
It is worth emphasizing that our method adds noise to original data, independent of specific optimization methods, which means that our proposed method is a general paradigm.
Moreover, detailed analysis of the privacy preserved on individuals of our method and how to improve the privacy of individuals will also be paid attentions in future work.


\bibliographystyle{ACM-Reference-Format}
\bibliography{ref}


\begin{thebibliography}{50}


\ifx \showCODEN    \undefined \def \showCODEN     #1{\unskip}     \fi
\ifx \showDOI      \undefined \def \showDOI       #1{#1}\fi
\ifx \showISBNx    \undefined \def \showISBNx     #1{\unskip}     \fi
\ifx \showISBNxiii \undefined \def \showISBNxiii  #1{\unskip}     \fi
\ifx \showISSN     \undefined \def \showISSN      #1{\unskip}     \fi
\ifx \showLCCN     \undefined \def \showLCCN      #1{\unskip}     \fi
\ifx \shownote     \undefined \def \shownote      #1{#1}          \fi
\ifx \showarticletitle \undefined \def \showarticletitle #1{#1}   \fi
\ifx \showURL      \undefined \def \showURL       {\relax}        \fi
\providecommand\bibfield[2]{#2}
\providecommand\bibinfo[2]{#2}
\providecommand\natexlab[1]{#1}
\providecommand\showeprint[2][]{arXiv:#2}

\bibitem[\protect\citeauthoryear{Abadi, Chu, Goodfellow, McMahan, Mironov,
  Talwar, and Zhang}{Abadi et~al\mbox{.}}{2016}]%
        {15}
\bibfield{author}{\bibinfo{person}{Martin Abadi}, \bibinfo{person}{Andy Chu},
  \bibinfo{person}{Ian Goodfellow}, \bibinfo{person}{H~Brendan McMahan},
  \bibinfo{person}{Ilya Mironov}, \bibinfo{person}{Kunal Talwar}, {and}
  \bibinfo{person}{Li Zhang}.} \bibinfo{year}{2016}\natexlab{}.
\newblock \showarticletitle{Deep learning with differential privacy}. In
  \bibinfo{booktitle}{\emph{Proceedings of the 2016 ACM SIGSAC Conference on
  Computer and Communications Security}}. \bibinfo{pages}{308--318}.
\newblock


\bibitem[\protect\citeauthoryear{Agrawal and Srikant}{Agrawal and
  Srikant}{2000}]%
        {49}
\bibfield{author}{\bibinfo{person}{Rakesh Agrawal} {and}
  \bibinfo{person}{Ramakrishnan Srikant}.} \bibinfo{year}{2000}\natexlab{}.
\newblock \showarticletitle{Privacy-preserving data mining}. In
  \bibinfo{booktitle}{\emph{ACM Sigmod Record}}, Vol.~\bibinfo{volume}{29}.
  \bibinfo{pages}{439--450}.
\newblock


\bibitem[\protect\citeauthoryear{Arora and Upadhyay}{Arora and
  Upadhyay}{2019}]%
        {56}
\bibfield{author}{\bibinfo{person}{Raman Arora} {and} \bibinfo{person}{Jalaj
  Upadhyay}.} \bibinfo{year}{2019}\natexlab{}.
\newblock \showarticletitle{On Differentially Private Graph Sparsification and
  Applications}.
\newblock In \bibinfo{booktitle}{\emph{Advances in Neural Information
  Processing Systems 32}}, \bibfield{editor}{\bibinfo{person}{H.~Wallach},
  \bibinfo{person}{H.~Larochelle}, \bibinfo{person}{A.~Beygelzimer},
  \bibinfo{person}{F.~d\textquotesingle Alch\'{e}-Buc},
  \bibinfo{person}{E.~Fox}, {and} \bibinfo{person}{R.~Garnett}} (Eds.).
  \bibinfo{publisher}{Curran Associates, Inc.}, \bibinfo{pages}{13378--13389}.
\newblock
\urldef\tempurl%
\url{http://papers.nips.cc/paper/9494-on-differentially-private-graph-sparsification-and-applications.pdf}
\showURL{%
\tempurl}


\bibitem[\protect\citeauthoryear{Bassily, Smith, and Thakurta}{Bassily
  et~al\mbox{.}}{2014}]%
        {21}
\bibfield{author}{\bibinfo{person}{Raef Bassily}, \bibinfo{person}{Adam Smith},
  {and} \bibinfo{person}{Abhradeep Thakurta}.} \bibinfo{year}{2014}\natexlab{}.
\newblock \showarticletitle{Private empirical risk minimization: Efficient
  algorithms and tight error bounds}. In \bibinfo{booktitle}{\emph{2014 IEEE
  55th Annual Symposium on Foundations of Computer Science}}.
  \bibinfo{pages}{464--473}.
\newblock


\bibitem[\protect\citeauthoryear{Beimel, Nissim, and Omri}{Beimel
  et~al\mbox{.}}{2008}]%
        {45}
\bibfield{author}{\bibinfo{person}{Amos Beimel}, \bibinfo{person}{Kobbi
  Nissim}, {and} \bibinfo{person}{Eran Omri}.} \bibinfo{year}{2008}\natexlab{}.
\newblock \showarticletitle{Distributed private data analysis: Simultaneously
  solving how and what}. In \bibinfo{booktitle}{\emph{Annual International
  Cryptology Conference}}. Springer, \bibinfo{pages}{451--468}.
\newblock


\bibitem[\protect\citeauthoryear{Bernstein and Sheldon}{Bernstein and
  Sheldon}{2019}]%
        {51}
\bibfield{author}{\bibinfo{person}{Garrett Bernstein} {and}
  \bibinfo{person}{Daniel~R Sheldon}.} \bibinfo{year}{2019}\natexlab{}.
\newblock \showarticletitle{Differentially Private Bayesian Linear Regression}.
\newblock In \bibinfo{booktitle}{\emph{Advances in Neural Information
  Processing Systems 32}}, \bibfield{editor}{\bibinfo{person}{H.~Wallach},
  \bibinfo{person}{H.~Larochelle}, \bibinfo{person}{A.~Beygelzimer},
  \bibinfo{person}{F.~d\textquotesingle Alch\'{e}-Buc},
  \bibinfo{person}{E.~Fox}, {and} \bibinfo{person}{R.~Garnett}} (Eds.).
  \bibinfo{publisher}{Curran Associates, Inc.}, \bibinfo{pages}{523--533}.
\newblock
\urldef\tempurl%
\url{http://papers.nips.cc/paper/8343-differentially-private-bayesian-linear-regression.pdf}
\showURL{%
\tempurl}


\bibitem[\protect\citeauthoryear{Bontempi and Worldline}{Bontempi and
  Worldline}{2018}]%
        {35}
\bibfield{author}{\bibinfo{person}{Gianluca Bontempi} {and}
  \bibinfo{person}{Worldline}.} \bibinfo{year}{2018}\natexlab{}.
\newblock \bibinfo{title}{{ULB} The Machine Learning Group}.
\newblock
\newblock
\urldef\tempurl%
\url{http://mlg.ulb.ac.be}
\showURL{%
\tempurl}


\bibitem[\protect\citeauthoryear{Bulgarevich, Tsukamoto, Kasuya, Demura, and
  Watanabe}{Bulgarevich et~al\mbox{.}}{2018}]%
        {3}
\bibfield{author}{\bibinfo{person}{Dmitry~S Bulgarevich},
  \bibinfo{person}{Susumu Tsukamoto}, \bibinfo{person}{Tadashi Kasuya},
  \bibinfo{person}{Masahiko Demura}, {and} \bibinfo{person}{Makoto Watanabe}.}
  \bibinfo{year}{2018}\natexlab{}.
\newblock \showarticletitle{Pattern recognition with machine learning on
  optical microscopy images of typical metallurgical microstructures}.
\newblock \bibinfo{journal}{\emph{Scientific reports}} \bibinfo{volume}{8},
  \bibinfo{number}{1} (\bibinfo{year}{2018}), \bibinfo{pages}{2078}.
\newblock


\bibitem[\protect\citeauthoryear{Bun and Steinke}{Bun and Steinke}{2016}]%
        {28}
\bibfield{author}{\bibinfo{person}{Mark Bun} {and} \bibinfo{person}{Thomas
  Steinke}.} \bibinfo{year}{2016}\natexlab{}.
\newblock \showarticletitle{Concentrated differential privacy: Simplifications,
  extensions, and lower bounds}. In \bibinfo{booktitle}{\emph{Theory of
  Cryptography Conference}}. Springer, \bibinfo{pages}{635--658}.
\newblock


\bibitem[\protect\citeauthoryear{Chaudhuri and Monteleoni}{Chaudhuri and
  Monteleoni}{2009}]%
        {9}
\bibfield{author}{\bibinfo{person}{Kamalika Chaudhuri} {and}
  \bibinfo{person}{Claire Monteleoni}.} \bibinfo{year}{2009}\natexlab{}.
\newblock \showarticletitle{Privacy-preserving logistic regression}. In
  \bibinfo{booktitle}{\emph{Advances in neural information processing
  systems}}. \bibinfo{pages}{289--296}.
\newblock


\bibitem[\protect\citeauthoryear{Chaudhuri, Monteleoni, and Sarwate}{Chaudhuri
  et~al\mbox{.}}{2011}]%
        {18}
\bibfield{author}{\bibinfo{person}{Kamalika Chaudhuri}, \bibinfo{person}{Claire
  Monteleoni}, {and} \bibinfo{person}{Anand~D Sarwate}.}
  \bibinfo{year}{2011}\natexlab{}.
\newblock \showarticletitle{Differentially private empirical risk
  minimization}.
\newblock \bibinfo{journal}{\emph{Journal of Machine Learning Research}}
  \bibinfo{volume}{12}, \bibinfo{number}{Mar} (\bibinfo{year}{2011}),
  \bibinfo{pages}{1069--1109}.
\newblock


\bibitem[\protect\citeauthoryear{Chaudhuri, Sarwate, and Sinha}{Chaudhuri
  et~al\mbox{.}}{2013}]%
        {11}
\bibfield{author}{\bibinfo{person}{Kamalika Chaudhuri},
  \bibinfo{person}{Anand~D Sarwate}, {and} \bibinfo{person}{Kaushik Sinha}.}
  \bibinfo{year}{2013}\natexlab{}.
\newblock \showarticletitle{A near-optimal algorithm for differentially-private
  principal components}.
\newblock \bibinfo{journal}{\emph{The Journal of Machine Learning Research}}
  \bibinfo{volume}{14}, \bibinfo{number}{1} (\bibinfo{year}{2013}),
  \bibinfo{pages}{2905--2943}.
\newblock


\bibitem[\protect\citeauthoryear{Csiba and Richt{\'a}rik}{Csiba and
  Richt{\'a}rik}{2017}]%
        {29}
\bibfield{author}{\bibinfo{person}{Dominik Csiba} {and} \bibinfo{person}{Peter
  Richt{\'a}rik}.} \bibinfo{year}{2017}\natexlab{}.
\newblock \showarticletitle{Global Convergence of Arbitrary-Block Gradient
  Methods for Generalized Polyak-$\{$$\backslash$L$\}$ ojasiewicz Functions}.
\newblock \bibinfo{journal}{\emph{arXiv preprint arXiv:1709.03014}}
  (\bibinfo{year}{2017}).
\newblock


\bibitem[\protect\citeauthoryear{Dua and Graff}{Dua and Graff}{2017}]%
        {32}
\bibfield{author}{\bibinfo{person}{Dheeru Dua} {and} \bibinfo{person}{Casey
  Graff}.} \bibinfo{year}{2017}\natexlab{}.
\newblock \bibinfo{title}{{UCI} Machine Learning Repository}.
\newblock
\newblock
\urldef\tempurl%
\url{http://archive.ics.uci.edu/ml}
\showURL{%
\tempurl}


\bibitem[\protect\citeauthoryear{Duchi, Jordan, and Wainwright}{Duchi
  et~al\mbox{.}}{2013}]%
        {48}
\bibfield{author}{\bibinfo{person}{John~C Duchi}, \bibinfo{person}{Michael~I
  Jordan}, {and} \bibinfo{person}{Martin~J Wainwright}.}
  \bibinfo{year}{2013}\natexlab{}.
\newblock \showarticletitle{Local privacy and statistical minimax rates}. In
  \bibinfo{booktitle}{\emph{2013 IEEE 54th Annual Symposium on Foundations of
  Computer Science}}. \bibinfo{pages}{429--438}.
\newblock


\bibitem[\protect\citeauthoryear{Duchi, Jordan, and Wainwright}{Duchi
  et~al\mbox{.}}{2018}]%
        {59}
\bibfield{author}{\bibinfo{person}{John~C Duchi}, \bibinfo{person}{Michael~I
  Jordan}, {and} \bibinfo{person}{Martin~J Wainwright}.}
  \bibinfo{year}{2018}\natexlab{}.
\newblock \showarticletitle{Minimax optimal procedures for locally private
  estimation}.
\newblock \bibinfo{journal}{\emph{J. Amer. Statist. Assoc.}}
  \bibinfo{volume}{113}, \bibinfo{number}{521} (\bibinfo{year}{2018}),
  \bibinfo{pages}{182--201}.
\newblock


\bibitem[\protect\citeauthoryear{Dwork}{Dwork}{2011}]%
        {7}
\bibfield{author}{\bibinfo{person}{Cynthia Dwork}.}
  \bibinfo{year}{2011}\natexlab{}.
\newblock \showarticletitle{Differential privacy}.
\newblock \bibinfo{journal}{\emph{Encyclopedia of Cryptography and Security}}
  (\bibinfo{year}{2011}), \bibinfo{pages}{338--340}.
\newblock


\bibitem[\protect\citeauthoryear{Dwork, McSherry, Nissim, and Smith}{Dwork
  et~al\mbox{.}}{2006}]%
        {8}
\bibfield{author}{\bibinfo{person}{Cynthia Dwork}, \bibinfo{person}{Frank
  McSherry}, \bibinfo{person}{Kobbi Nissim}, {and} \bibinfo{person}{Adam
  Smith}.} \bibinfo{year}{2006}\natexlab{}.
\newblock \showarticletitle{Calibrating noise to sensitivity in private data
  analysis}. In \bibinfo{booktitle}{\emph{Theory of cryptography conference}}.
  Springer, \bibinfo{pages}{265--284}.
\newblock


\bibitem[\protect\citeauthoryear{Dwork, Roth, et~al\mbox{.}}{Dwork
  et~al\mbox{.}}{2014}]%
        {27}
\bibfield{author}{\bibinfo{person}{Cynthia Dwork}, \bibinfo{person}{Aaron
  Roth}, {et~al\mbox{.}}} \bibinfo{year}{2014}\natexlab{}.
\newblock \showarticletitle{The algorithmic foundations of differential
  privacy}.
\newblock \bibinfo{journal}{\emph{Foundations and Trends{\textregistered} in
  Theoretical Computer Science}} \bibinfo{volume}{9}, \bibinfo{number}{3--4}
  (\bibinfo{year}{2014}), \bibinfo{pages}{211--407}.
\newblock


\bibitem[\protect\citeauthoryear{Dwork, Rothblum, and Vadhan}{Dwork
  et~al\mbox{.}}{2010}]%
        {16}
\bibfield{author}{\bibinfo{person}{Cynthia Dwork}, \bibinfo{person}{Guy~N
  Rothblum}, {and} \bibinfo{person}{Salil Vadhan}.}
  \bibinfo{year}{2010}\natexlab{}.
\newblock \showarticletitle{Boosting and differential privacy}. In
  \bibinfo{booktitle}{\emph{2010 IEEE 51st Annual Symposium on Foundations of
  Computer Science}}. \bibinfo{pages}{51--60}.
\newblock


\bibitem[\protect\citeauthoryear{Fan}{Fan}{2018}]%
        {44}
\bibfield{author}{\bibinfo{person}{Liyue Fan}.}
  \bibinfo{year}{2018}\natexlab{}.
\newblock \showarticletitle{Image pixelization with differential privacy}. In
  \bibinfo{booktitle}{\emph{IFIP Annual Conference on Data and Applications
  Security and Privacy}}. Springer, \bibinfo{pages}{148--162}.
\newblock


\bibitem[\protect\citeauthoryear{Fredrikson, Lantz, Jha, Lin, Page, and
  Ristenpart}{Fredrikson et~al\mbox{.}}{2014}]%
        {5}
\bibfield{author}{\bibinfo{person}{Matthew Fredrikson}, \bibinfo{person}{Eric
  Lantz}, \bibinfo{person}{Somesh Jha}, \bibinfo{person}{Simon Lin},
  \bibinfo{person}{David Page}, {and} \bibinfo{person}{Thomas Ristenpart}.}
  \bibinfo{year}{2014}\natexlab{}.
\newblock \showarticletitle{Privacy in pharmacogenetics: An end-to-end case
  study of personalized warfarin dosing}. In \bibinfo{booktitle}{\emph{23rd
  $\{$USENIX$\}$ Security Symposium ($\{$USENIX$\}$ Security 14)}}.
  \bibinfo{pages}{17--32}.
\newblock


\bibitem[\protect\citeauthoryear{Fu, Levin-Schwartz, Lin, and Zhang}{Fu
  et~al\mbox{.}}{2019}]%
        {4}
\bibfield{author}{\bibinfo{person}{Geng-Shen Fu}, \bibinfo{person}{Yuri
  Levin-Schwartz}, \bibinfo{person}{Qiu-Hua Lin}, {and} \bibinfo{person}{Da
  Zhang}.} \bibinfo{year}{2019}\natexlab{}.
\newblock \showarticletitle{Machine Learning for Medical Imaging}.
\newblock \bibinfo{journal}{\emph{Journal of healthcare engineering}}
  \bibinfo{volume}{2019} (\bibinfo{year}{2019}).
\newblock


\bibitem[\protect\citeauthoryear{Fukuchi, Tran, and Sakuma}{Fukuchi
  et~al\mbox{.}}{2017}]%
        {62}
\bibfield{author}{\bibinfo{person}{Kazuto Fukuchi}, \bibinfo{person}{Quang~Khai
  Tran}, {and} \bibinfo{person}{Jun Sakuma}.} \bibinfo{year}{2017}\natexlab{}.
\newblock \showarticletitle{Differentially Private Empirical Risk Minimization
  with Input Perturbation}. In \bibinfo{booktitle}{\emph{International
  Conference on Discovery Science}}. Springer, \bibinfo{pages}{82--90}.
\newblock


\bibitem[\protect\citeauthoryear{Hettich and Bay}{Hettich and Bay}{1999}]%
        {31}
\bibfield{author}{\bibinfo{person}{S. Hettich} {and} \bibinfo{person}{S.~D.
  Bay}.} \bibinfo{year}{1999}\natexlab{}.
\newblock \bibinfo{title}{The UCI KDD Archive [http://kdd.ics.uci.edu].}
\newblock
\newblock


\bibitem[\protect\citeauthoryear{Hill, Zhou, Saul, and Shacham}{Hill
  et~al\mbox{.}}{2016}]%
        {58}
\bibfield{author}{\bibinfo{person}{Steven Hill}, \bibinfo{person}{Zhimin Zhou},
  \bibinfo{person}{Lawrence Saul}, {and} \bibinfo{person}{Hovav Shacham}.}
  \bibinfo{year}{2016}\natexlab{}.
\newblock \showarticletitle{On the (in) effectiveness of mosaicing and blurring
  as tools for document redaction}.
\newblock \bibinfo{journal}{\emph{Proceedings on Privacy Enhancing
  Technologies}} \bibinfo{volume}{2016}, \bibinfo{number}{4}
  (\bibinfo{year}{2016}), \bibinfo{pages}{403--417}.
\newblock


\bibitem[\protect\citeauthoryear{Kairouz, Oh, and Viswanath}{Kairouz
  et~al\mbox{.}}{2014}]%
        {40}
\bibfield{author}{\bibinfo{person}{Peter Kairouz}, \bibinfo{person}{Sewoong
  Oh}, {and} \bibinfo{person}{Pramod Viswanath}.}
  \bibinfo{year}{2014}\natexlab{}.
\newblock \showarticletitle{Extremal mechanisms for local differential
  privacy}. In \bibinfo{booktitle}{\emph{Advances in neural information
  processing systems}}. \bibinfo{pages}{2879--2887}.
\newblock


\bibitem[\protect\citeauthoryear{Karimi, Nutini, and Schmidt}{Karimi
  et~al\mbox{.}}{2016}]%
        {30}
\bibfield{author}{\bibinfo{person}{Hamed Karimi}, \bibinfo{person}{Julie
  Nutini}, {and} \bibinfo{person}{Mark Schmidt}.}
  \bibinfo{year}{2016}\natexlab{}.
\newblock \showarticletitle{Linear convergence of gradient and
  proximal-gradient methods under the polyak-{\l}ojasiewicz condition}. In
  \bibinfo{booktitle}{\emph{Joint European Conference on Machine Learning and
  Knowledge Discovery in Databases}}. Springer, \bibinfo{pages}{795--811}.
\newblock


\bibitem[\protect\citeauthoryear{Kifer, Smith, and Thakurta}{Kifer
  et~al\mbox{.}}{2012}]%
        {23}
\bibfield{author}{\bibinfo{person}{Daniel Kifer}, \bibinfo{person}{Adam Smith},
  {and} \bibinfo{person}{Abhradeep Thakurta}.} \bibinfo{year}{2012}\natexlab{}.
\newblock \showarticletitle{Private convex empirical risk minimization and
  high-dimensional regression}. In \bibinfo{booktitle}{\emph{Conference on
  Learning Theory}}. \bibinfo{pages}{25--1}.
\newblock


\bibitem[\protect\citeauthoryear{Lee, Kim, Lee, Suh, and Ramchandran}{Lee
  et~al\mbox{.}}{2019}]%
        {63}
\bibfield{author}{\bibinfo{person}{Kangwook Lee}, \bibinfo{person}{Hoon Kim},
  \bibinfo{person}{Kyungmin Lee}, \bibinfo{person}{Changho Suh}, {and}
  \bibinfo{person}{Kannan Ramchandran}.} \bibinfo{year}{2019}\natexlab{}.
\newblock \showarticletitle{Synthesizing Differentially Private Datasets using
  Random Mixing}. In \bibinfo{booktitle}{\emph{2019 IEEE International
  Symposium on Information Theory (ISIT)}}. \bibinfo{pages}{542--546}.
\newblock


\bibitem[\protect\citeauthoryear{LeTien, Habrard, and Sebban}{LeTien
  et~al\mbox{.}}{2019}]%
        {13}
\bibfield{author}{\bibinfo{person}{Nam LeTien}, \bibinfo{person}{Amaury
  Habrard}, {and} \bibinfo{person}{Marc Sebban}.}
  \bibinfo{year}{2019}\natexlab{}.
\newblock \showarticletitle{Differentially Private Optimal Transport:
  Application to Domain Adaptation}. In \bibinfo{booktitle}{\emph{Proceedings
  of the Twenty-Eighth International Joint Conference on Artificial
  Intelligence, {IJCAI-19}}}. \bibinfo{publisher}{International Joint
  Conferences on Artificial Intelligence Organization},
  \bibinfo{pages}{2852--2858}.
\newblock
\urldef\tempurl%
\url{https://doi.org/10.24963/ijcai.2019/395}
\showDOI{\tempurl}


\bibitem[\protect\citeauthoryear{Mangasarian and Wolberg}{Mangasarian and
  Wolberg}{1990}]%
        {34}
\bibfield{author}{\bibinfo{person}{Olvi~L Mangasarian} {and}
  \bibinfo{person}{William~H Wolberg}.} \bibinfo{year}{1990}\natexlab{}.
\newblock \bibinfo{booktitle}{\emph{Cancer diagnosis via linear programming}}.
\newblock \bibinfo{type}{{T}echnical {R}eport}.
  \bibinfo{institution}{University of Wisconsin-Madison Department of Computer
  Sciences}.
\newblock


\bibitem[\protect\citeauthoryear{Moro, Cortez, and Rita}{Moro
  et~al\mbox{.}}{2014}]%
        {33}
\bibfield{author}{\bibinfo{person}{S{\'e}rgio Moro}, \bibinfo{person}{Paulo
  Cortez}, {and} \bibinfo{person}{Paulo Rita}.}
  \bibinfo{year}{2014}\natexlab{}.
\newblock \showarticletitle{A data-driven approach to predict the success of
  bank telemarketing}.
\newblock \bibinfo{journal}{\emph{Decision Support Systems}}
  \bibinfo{volume}{62} (\bibinfo{year}{2014}), \bibinfo{pages}{22--31}.
\newblock


\bibitem[\protect\citeauthoryear{Rahman, Janmohamed, Pirbaglou, Clarke, Ritvo,
  Heffernan, and Katz}{Rahman et~al\mbox{.}}{2018}]%
        {2}
\bibfield{author}{\bibinfo{person}{Quazi~Abidur Rahman}, \bibinfo{person}{Tahir
  Janmohamed}, \bibinfo{person}{Meysam Pirbaglou}, \bibinfo{person}{Hance
  Clarke}, \bibinfo{person}{Paul Ritvo}, \bibinfo{person}{Jane~M Heffernan},
  {and} \bibinfo{person}{Joel Katz}.} \bibinfo{year}{2018}\natexlab{}.
\newblock \showarticletitle{Defining and Predicting Pain Volatility in Users of
  the Manage My Pain App: Analysis Using Data Mining and Machine Learning
  Methods}.
\newblock \bibinfo{journal}{\emph{Journal of medical Internet research}}
  \bibinfo{volume}{20}, \bibinfo{number}{11} (\bibinfo{year}{2018}),
  \bibinfo{pages}{e12001}.
\newblock


\bibitem[\protect\citeauthoryear{Sealfon and Ullman}{Sealfon and
  Ullman}{2019}]%
        {25}
\bibfield{author}{\bibinfo{person}{Adam Sealfon} {and}
  \bibinfo{person}{Jonathan Ullman}.} \bibinfo{year}{2019}\natexlab{}.
\newblock \showarticletitle{Efficiently Estimating Erdos-Renyi Graphs with Node
  Differential Privacy}.
\newblock \bibinfo{journal}{\emph{arXiv preprint arXiv:1905.10477}}
  (\bibinfo{year}{2019}).
\newblock


\bibitem[\protect\citeauthoryear{Shokri and Shmatikov}{Shokri and
  Shmatikov}{2015}]%
        {14}
\bibfield{author}{\bibinfo{person}{Reza Shokri} {and} \bibinfo{person}{Vitaly
  Shmatikov}.} \bibinfo{year}{2015}\natexlab{}.
\newblock \showarticletitle{Privacy-preserving deep learning}. In
  \bibinfo{booktitle}{\emph{Proceedings of the 22nd ACM SIGSAC conference on
  computer and communications security}}. \bibinfo{pages}{1310--1321}.
\newblock


\bibitem[\protect\citeauthoryear{Shokri, Stronati, Song, and Shmatikov}{Shokri
  et~al\mbox{.}}{2017}]%
        {6}
\bibfield{author}{\bibinfo{person}{Reza Shokri}, \bibinfo{person}{Marco
  Stronati}, \bibinfo{person}{Congzheng Song}, {and} \bibinfo{person}{Vitaly
  Shmatikov}.} \bibinfo{year}{2017}\natexlab{}.
\newblock \showarticletitle{Membership inference attacks against machine
  learning models}. In \bibinfo{booktitle}{\emph{2017 IEEE Symposium on
  Security and Privacy (SP)}}. \bibinfo{pages}{3--18}.
\newblock


\bibitem[\protect\citeauthoryear{Smith, Lopez, Zwiessele, and Lawrence}{Smith
  et~al\mbox{.}}{2018}]%
        {10}
\bibfield{author}{\bibinfo{person}{M Smith}, \bibinfo{person}{Alvarez Lopez},
  \bibinfo{person}{M Zwiessele}, {and} \bibinfo{person}{N Lawrence}.}
  \bibinfo{year}{2018}\natexlab{}.
\newblock \showarticletitle{Differentially Private Regression using Gaussian
  Processes}. In \bibinfo{booktitle}{\emph{Proceedings of Machine Learning
  Research}}, Vol.~\bibinfo{volume}{84}.
\newblock


\bibitem[\protect\citeauthoryear{Ullman and Sealfon}{Ullman and
  Sealfon}{2019}]%
        {52}
\bibfield{author}{\bibinfo{person}{Jonathan Ullman} {and} \bibinfo{person}{Adam
  Sealfon}.} \bibinfo{year}{2019}\natexlab{}.
\newblock \showarticletitle{Efficiently Estimating Erdos-Renyi Graphs with Node
  Differential Privacy}.
\newblock In \bibinfo{booktitle}{\emph{Advances in Neural Information
  Processing Systems 32}}, \bibfield{editor}{\bibinfo{person}{H.~Wallach},
  \bibinfo{person}{H.~Larochelle}, \bibinfo{person}{A.~Beygelzimer},
  \bibinfo{person}{F.~d\textquotesingle Alch\'{e}-Buc},
  \bibinfo{person}{E.~Fox}, {and} \bibinfo{person}{R.~Garnett}} (Eds.).
  \bibinfo{publisher}{Curran Associates, Inc.}, \bibinfo{pages}{3765--3775}.
\newblock
\urldef\tempurl%
\url{http://papers.nips.cc/paper/8633-efficiently-estimating-erdos-renyi-graphs-with-node-differential-privacy.pdf}
\showURL{%
\tempurl}


\bibitem[\protect\citeauthoryear{Wang, Gaboardi, and Xu}{Wang
  et~al\mbox{.}}{2018}]%
        {43}
\bibfield{author}{\bibinfo{person}{Di Wang}, \bibinfo{person}{Marco Gaboardi},
  {and} \bibinfo{person}{Jinhui Xu}.} \bibinfo{year}{2018}\natexlab{}.
\newblock \showarticletitle{Empirical risk minimization in non-interactive
  local differential privacy revisited}. In \bibinfo{booktitle}{\emph{Advances
  in Neural Information Processing Systems}}. \bibinfo{pages}{965--974}.
\newblock


\bibitem[\protect\citeauthoryear{Wang, Smith, and Xu}{Wang
  et~al\mbox{.}}{2019a}]%
        {46}
\bibfield{author}{\bibinfo{person}{Di Wang}, \bibinfo{person}{Adam Smith},
  {and} \bibinfo{person}{Jinhui Xu}.} \bibinfo{year}{2019}\natexlab{a}.
\newblock \showarticletitle{Noninteractive locally private learning of linear
  models via polynomial approximations}. In
  \bibinfo{booktitle}{\emph{Algorithmic Learning Theory}}.
  \bibinfo{pages}{897--902}.
\newblock


\bibitem[\protect\citeauthoryear{Wang and Xu}{Wang and Xu}{2019}]%
        {12}
\bibfield{author}{\bibinfo{person}{Di Wang} {and} \bibinfo{person}{Jinhui Xu}.}
  \bibinfo{year}{2019}\natexlab{}.
\newblock \showarticletitle{Principal Component Analysis in the Local
  Differential Privacy Model}. In \bibinfo{booktitle}{\emph{Proceedings of the
  Twenty-Eighth International Joint Conference on Artificial Intelligence,
  {IJCAI-19}}}. \bibinfo{publisher}{International Joint Conferences on
  Artificial Intelligence Organization}, \bibinfo{pages}{4795--4801}.
\newblock
\urldef\tempurl%
\url{https://doi.org/10.24963/ijcai.2019/666}
\showDOI{\tempurl}


\bibitem[\protect\citeauthoryear{Wang, Ye, and Xu}{Wang et~al\mbox{.}}{2017}]%
        {20}
\bibfield{author}{\bibinfo{person}{Di Wang}, \bibinfo{person}{Minwei Ye}, {and}
  \bibinfo{person}{Jinhui Xu}.} \bibinfo{year}{2017}\natexlab{}.
\newblock \showarticletitle{Differentially private empirical risk minimization
  revisited: Faster and more general}. In \bibinfo{booktitle}{\emph{Advances in
  Neural Information Processing Systems}}. \bibinfo{pages}{2722--2731}.
\newblock


\bibitem[\protect\citeauthoryear{Wang, Xiao, Yang, Zhao, Hui, Shin, Shin, and
  Yu}{Wang et~al\mbox{.}}{2019b}]%
        {41}
\bibfield{author}{\bibinfo{person}{Ning Wang}, \bibinfo{person}{Xiaokui Xiao},
  \bibinfo{person}{Yin Yang}, \bibinfo{person}{Jun Zhao},
  \bibinfo{person}{Siu~Cheung Hui}, \bibinfo{person}{Hyejin Shin},
  \bibinfo{person}{Junbum Shin}, {and} \bibinfo{person}{Ge Yu}.}
  \bibinfo{year}{2019}\natexlab{b}.
\newblock \showarticletitle{Collecting and Analyzing Multidimensional Data with
  Local Differential Privacy}. In \bibinfo{booktitle}{\emph{2019 IEEE 35th
  International Conference on Data Engineering (ICDE)}}.
  \bibinfo{pages}{638--649}.
\newblock


\bibitem[\protect\citeauthoryear{Wu, Zhao, Xu, Chen, Wang, Zhang, Sun, and
  Zhou}{Wu et~al\mbox{.}}{2019}]%
        {24}
\bibfield{author}{\bibinfo{person}{Bingzhe Wu}, \bibinfo{person}{Shiwan Zhao},
  \bibinfo{person}{Haoyang Xu}, \bibinfo{person}{ChaoChao Chen},
  \bibinfo{person}{Li Wang}, \bibinfo{person}{Xiaolu Zhang},
  \bibinfo{person}{Guangyu Sun}, {and} \bibinfo{person}{Jun Zhou}.}
  \bibinfo{year}{2019}\natexlab{}.
\newblock \showarticletitle{Generalization in Generative Adversarial Networks:
  A Novel Perspective from Privacy Protection}.
\newblock \bibinfo{journal}{\emph{arXiv preprint arXiv:1908.07882}}
  (\bibinfo{year}{2019}).
\newblock


\bibitem[\protect\citeauthoryear{Xiao and Zhang}{Xiao and Zhang}{2014}]%
        {26}
\bibfield{author}{\bibinfo{person}{Lin Xiao} {and} \bibinfo{person}{Tong
  Zhang}.} \bibinfo{year}{2014}\natexlab{}.
\newblock \showarticletitle{A proximal stochastic gradient method with
  progressive variance reduction}.
\newblock \bibinfo{journal}{\emph{SIAM Journal on Optimization}}
  \bibinfo{volume}{24}, \bibinfo{number}{4} (\bibinfo{year}{2014}),
  \bibinfo{pages}{2057--2075}.
\newblock


\bibitem[\protect\citeauthoryear{Xu, Ren, Zhang, Zhang, Qin, and Ren}{Xu
  et~al\mbox{.}}{2019}]%
        {36}
\bibfield{author}{\bibinfo{person}{Chugui Xu}, \bibinfo{person}{Ju Ren},
  \bibinfo{person}{Deyu Zhang}, \bibinfo{person}{Yaoxue Zhang},
  \bibinfo{person}{Zhan Qin}, {and} \bibinfo{person}{Kui Ren}.}
  \bibinfo{year}{2019}\natexlab{}.
\newblock \showarticletitle{GANobfuscator: Mitigating information leakage under
  GAN via differential privacy}.
\newblock \bibinfo{journal}{\emph{IEEE Transactions on Information Forensics
  and Security}} \bibinfo{volume}{14}, \bibinfo{number}{9}
  (\bibinfo{year}{2019}), \bibinfo{pages}{2358--2371}.
\newblock


\bibitem[\protect\citeauthoryear{Yee, Sagadevan, and Malim}{Yee
  et~al\mbox{.}}{2018}]%
        {1}
\bibfield{author}{\bibinfo{person}{Ong~Shu Yee}, \bibinfo{person}{Saravanan
  Sagadevan}, {and} \bibinfo{person}{Nurul Hashimah Ahamed~Hassain Malim}.}
  \bibinfo{year}{2018}\natexlab{}.
\newblock \showarticletitle{Credit card fraud detection using machine learning
  as data mining technique}.
\newblock \bibinfo{journal}{\emph{Journal of Telecommunication, Electronic and
  Computer Engineering (JTEC)}} \bibinfo{volume}{10}, \bibinfo{number}{1-4}
  (\bibinfo{year}{2018}), \bibinfo{pages}{23--27}.
\newblock


\bibitem[\protect\citeauthoryear{Zhang, Zheng, Mou, and Wang}{Zhang
  et~al\mbox{.}}{2017}]%
        {22}
\bibfield{author}{\bibinfo{person}{Jiaqi Zhang}, \bibinfo{person}{Kai Zheng},
  \bibinfo{person}{Wenlong Mou}, {and} \bibinfo{person}{Liwei Wang}.}
  \bibinfo{year}{2017}\natexlab{}.
\newblock \showarticletitle{Efficient private ERM for smooth objectives}.
\newblock \bibinfo{journal}{\emph{arXiv preprint arXiv:1703.09947}}
  (\bibinfo{year}{2017}).
\newblock


\bibitem[\protect\citeauthoryear{Zhao, Ni, Hu, Chen, Zhou, Xiao, and Wu}{Zhao
  et~al\mbox{.}}{2018}]%
        {17}
\bibfield{author}{\bibinfo{person}{Lingchen Zhao}, \bibinfo{person}{Lihao Ni},
  \bibinfo{person}{Shengshan Hu}, \bibinfo{person}{Yaniiao Chen},
  \bibinfo{person}{Pan Zhou}, \bibinfo{person}{Fu Xiao}, {and}
  \bibinfo{person}{Libing Wu}.} \bibinfo{year}{2018}\natexlab{}.
\newblock \showarticletitle{InPrivate Digging: Enabling Tree-based Distributed
  Data Mining with Differential Privacy}. In \bibinfo{booktitle}{\emph{IEEE
  INFOCOM 2018-IEEE Conference on Computer Communications}}.
  \bibinfo{pages}{2087--2095}.
\newblock


\end{thebibliography}

\appendix

\section{Details of Proof}

\subsection{Theorem 1}
\begin{proof}
By observing that the noise added to data causes the perturbation on the gradient, we pay our attentions on the gradient descent descent process:
\begin{equation}\label{inputgradesc}
\theta_{t+1}=\theta_{t}-\alpha\nabla\hat{L}(\theta_{t})=\theta_{t}-\alpha\frac{1}{n}\sum_{i=1}^{n}y_i\nabla\ell(y_i\theta_{t}^{T}(x_i+z))(x_i+z),
\end{equation}
where $z\sim\mathcal{N}(0,\sigma^2)$ and $\alpha$ denotes the learning rate.

Then, considering about the $t^{th}$ query which may disclose privacy, the randomized mechanism $M_t$ is:
\begin{equation}\label{Mt}
M_t=\frac{1}{n}\sum_{i=1}^{n}y_i\nabla\ell(y_i\theta_{t}^{T}(x_i+z))(x_i+z).
\end{equation}

Denote probability distributions on adjacent databases $D$ and $D'$ over mechanism $M_t$ as $P$ and $Q$:
\begin{equation}\label{PQ}
\begin{aligned}
P&=\frac{1}{n}\sum_{i=1}^{n-1}y_i\nabla\ell(y_i\theta_{t}^{T}(x_i+z))(x_i+z)+\frac{1}{n}y_n\nabla\ell(y_n\theta_{t}^{T}(x_n+z))(x_n+z), \\
Q&=\frac{1}{n}\sum_{i=1}^{n-1}y_i\nabla\ell(y_i\theta_{t}^{T}(x_i+z))(x_i+z)+\frac{1}{n}y_n'\nabla\ell(y_n'\theta_{t}^{T}(x_n'+z))(x_n'+z),
\end{aligned}
\end{equation}
where we suppose that the single different data instance between $D$ and $D'$ is the $n^{th}$ one, denoted as $(x_n,y_n)$ and $(x_n',y_n')$, respectively.

For simplicity on expression, we set:
\begin{equation}\label{ABC}
\begin{aligned}
A&=\frac{1}{n}\sum_{i=1}^{n-1}y_i\nabla\ell(y_i\theta_{t}^{T}(x_i+z))x_i, B=\frac{1}{n}y_n\nabla\ell(y_n\theta_{t}^{T}(x_n+z))(x_n+z), \\
B'&=\frac{1}{n}y_n'\nabla\ell(y_n'\theta_{t}^{T}(x_n'+z))(x_n'+z), C=\frac{1}{n}\sum_{i=1}^{n-1}y_i\nabla\ell(y_i\theta_{t}^{T}(x_i+z)).
\end{aligned}
\end{equation}

Then, by (\ref{PQ}), (\ref{ABC}) and note that $z \sim \mathcal{N}(0,\sigma^2)$, we have:
\begin{equation}\label{simpPQ}
P=\mathcal{N}(A+B,C\sigma^2), \quad Q=\mathcal{N}(A+B',C\sigma^2).
\end{equation}

In moments accountant method proposed in \cite{15}, the $\lambda^{th}$ moment $\alpha_M(\lambda;D,D')$ on mechanism $M$ is defined as:
\begin{equation}\label{momentsaccountant}
\alpha_M(\lambda;D,D')=\log\mathbb{E}_{o\sim M(D)}\left[\exp(\lambda c(o;M,D,D'))\right],
\end{equation}
where $c(o;M,D,D')$ is privacy loss at the output $o$, defined as:
\begin{equation}\label{privacyloss}
c(o;M,D,D')=\log\frac{\mathbb{P}\left[M(D)=o\right]}{\mathbb{P}\left[M(D')=o\right]}.
\end{equation}

When it comes to privacy preserving, it is necessary to bound all possible $\alpha_M(\lambda;D,D')$, denoted as $\alpha_M(\lambda)$, which is defined as:
\begin{equation}\label{alphaM}
\alpha_M(\lambda)=\max_{D,D'}\alpha_M(\lambda;D,D').
\end{equation}

By Definition 2.1 in \cite{28}, $D_{\alpha}$ is defined as:
\begin{equation}\label{Dalpha}
D_{\alpha}(P\Vert Q)=\frac{1}{\alpha-1}\log\left(\mathbb{E}_{x\sim P}\left[\left(\frac{P(x)}{Q
(x)}\right)^{\alpha-1}\right]\right).
\end{equation}

By (\ref{momentsaccountant}), (\ref{privacyloss}), (\ref{alphaM}), (\ref{Dalpha}) and $P$, $Q$ in (\ref{simpPQ}), we have:
\begin{equation}\label{alphatoD}
\begin{aligned}
\alpha_{M_t}(\lambda)&=\log\mathbb{E}_{o\sim P}\left[\exp\left(\lambda\log(\frac{P}{Q})\right)\right]=\log\mathbb{E}_{o\sim P}\left[\left(\frac{P}{Q}\right)^\lambda\right] \\
&=\lambda D_{\lambda+1}(P\Vert Q).
\end{aligned}
\end{equation}

By (\ref{alphatoD}) and Lemma 2.5 in \cite{28}, we have:
\begin{equation}\label{alphaMtsimp}
\alpha_{M_t}(\lambda)=\lambda D_{\lambda+1}(P\Vert Q)=\frac{\lambda(\lambda+1)\|(A+B)-(A+B')\|^2}{2C\sigma^2}.
\end{equation}

By definitions of $B$ and $B'$ in (\ref{ABC}) and note that $\ell(\theta,x,y)$ is $G$-Lipschitz ($G$), we have:
\begin{equation}\label{B-B'}
\|B-B'\|=\|\frac{1}{n}\nabla\ell(\theta_{t},x_n+z,y_n)-\frac{1}{n}\nabla\ell(\theta_{t},x_n'+z,y_n')\|\overset{(G)}{\leq}\frac{2G}{n}.
\end{equation}

By \cite{29}, if function $\ell(\theta,x,y)$ is $\Delta$-strongly convex ($\Delta$), we have:
\begin{equation}\label{convexity}
\|\nabla\ell(\theta,x,y)\|^2 \geq 2\Delta(\ell(\theta,x,y)-\ell^{*}).
\end{equation}

Combining (\ref{convexity}) and the definition of $C$ in (\ref{ABC}), we have:
\begin{equation}\label{supC}
C=\frac{1}{n}\sum_{i=1}^{n-1}y_i\nabla\ell(y_i\theta_{t}^{T}(x_i+z))\overset{(\Delta)}{\geq}\frac{n-1}{n}\sqrt{2\Delta(\ell(\theta_t)-\ell^{*})}.
\end{equation}

In general, with the increasing of training iteration, loss of the model decreases.
i.e. $\ell(\theta_{t_1}) \leq \ell(\theta_{t_2})$ if $t_1 \geq t_2$.
So, we have:
\begin{equation}\label{supCT}
C \geq \frac{n-1}{n}\sqrt{2\Delta(\ell(\theta_T)-\ell^{*})}.
\end{equation}

Considering that $\ell(\theta_T)-\ell^{*}$ can be seemed as a constant, by (\ref{B-B'}) and (\ref{supCT}), for some constant $c_1$, (\ref{alphaMtsimp}) can be transferred to:
\begin{equation}\label{alphaMtbound}
\alpha_{M_t}(\lambda) \leq c_1\frac{\lambda(\lambda+1)G^2}{\sqrt{\Delta}\sigma^2n(n-1)}.
\end{equation}

By Theorem 2.1 in \cite{15}, we have:
\begin{equation}\label{boundoverT}
\alpha_M(\lambda) \leq \sum_{t=1}^{T}\alpha_{M_t}(\lambda).
\end{equation}

By summing over $T$ iterations on (\ref{alphaMtbound}), for some constant $c_2$:
\begin{equation}\label{alphaMbound}
\alpha_M(\lambda)\leq\sum_{t=1}^{T}\alpha_{M_t}(\lambda) \leq c_1\frac{\lambda(\lambda+1)G^2T}{\sqrt{\Delta}\sigma^2n(n-1)}\leq c_2\frac{\lambda^2G^2T}{\sqrt{\Delta}\sigma^2n(n-1)}.
\end{equation}

Taking $\sigma^2=c\frac{G^2T\log(1/\delta)}{n(n-1)\sqrt{\Delta}\epsilon^2}$ for some constant $c$, we can guarantee:
\begin{equation}\label{condition1}
\alpha_M(\lambda) \leq \frac{c_2\lambda^2G^2T}{\sigma^2n(n-1)\sqrt{\Delta}} \leq \frac{\lambda\epsilon}{2},
\end{equation}
and as a result, we have:
\begin{equation}\label{condtition2}
\delta \leq \exp(\frac{-\lambda\epsilon}{2}),
\end{equation}
leading $(\epsilon,\delta)$-differential privacy according to Theorem 2.2 in \cite{15}.
\end{proof}

\subsection{Theorem 2}
\begin{proof}
First, considering $\mathbb{E}\left[\hat{L}(\theta_{t+1})-\hat{L}(\theta_{t})\right]$ at round $t$:
\begin{equation}\label{boundatt}
\begin{aligned}
\mathbb{E}\left[\hat{L}(\theta_{t+1})-\hat{L}(\theta_{t})\right]=\mathbb{E}_{z}\left[\frac{1}{n}\sum_{i=1}^{n}\left[\ell(y_i\theta_{t+1}^{T}(x_i+z))-\ell(y_i\theta_{t}^{T}(x_i+z))\right]\right].
\end{aligned}
\end{equation}

Note that $\ell(\cdot)$ is $G$-Lipschitz ($G$), then for all $x,y$:
\begin{equation}\label{GLipschitz}
\ell(x)-\ell(y) \leq G\left|x-y\right|.
\end{equation}

By the combination of (\ref{boundatt}) and (\ref{GLipschitz}), without loss of generality:
\begin{equation}\label{transboundatt}
\begin{aligned}
\mathbb{E}\left[\hat{L}(\theta_{t+1})-\hat{L}(\theta_{t})\right]&\overset{(G)}{\leq}\frac{1}{n}\sum_{i=1}^{n}\mathbb{E}_{z}\left[G\left|y_i\theta_{t+1}^{T}(x_i+z)-y_i\theta_{t}^{T}(x_i+z)\right|\right] \\
&\leq\alpha G\frac{1}{n}\sum_{i=1}^{n}\mathbb{E}_{z}\left[y_i(-\nabla\hat{L}(\theta_{t}))x_i+y_i(-\nabla\hat{L}(\theta_{t}))z\right].
\end{aligned}
\end{equation}

By the definition of $\hat{L}(\theta)$ in (\ref{inputobj}), we have:
\begin{equation}\label{nablaLhat}
\nabla\hat{L}(\theta_{t})=\frac{1}{n}\sum_{i=1}^{n}y_i\nabla\ell(y_i\theta_{t}^{T}(x_i+z))(x_i+z).
\end{equation}

Note that $\nabla\ell(\cdot)$ is $L$-Lipschitz ($L$), then we have:
\begin{equation}\label{LLipschitz}
\nabla\ell(y_i\theta_{t}^{T}(x_i+z))-\nabla\ell(y_i\theta_{t}^{T}x_i)\geq -L\left|y_i\theta_{t}^{T}z\right|.
\end{equation}

Then, by (\ref{nablaLhat}) and (\ref{LLipschitz}), we have:
\begin{equation}\label{sup-nablaLhat}
\begin{aligned}
\nabla\hat{L}(\theta_{t})&=\frac{1}{n}\sum_{i=1}^{n}\left[y_i\nabla\ell(y_i\theta_{t}^{T}(x_i+z))(x_i+z)\right] \\
&\overset{(L)}{\geq}\frac{1}{n}\sum_{i=1}^{n}\left[y_i\left(\nabla\ell(y_i\theta_{t}^{T}x_i)-L\left|y_i\theta_{t}^{T}z\right|\right)(x_i+z)\right].
\end{aligned}
\end{equation}

Note that $y_i\in[-1,1]$ and $\|x_i\| \leq 1$, (\ref{sup-nablaLhat}) can be transferred to:
\begin{equation}\label{nablaLhatsimp}
\nabla\hat{L}(\theta_{t})\geq\nabla L(\theta_{t})-L\|\theta_{t}\|\|z\|-L\|\theta_{t}\|\|z\|^2+\frac{1}{n}\sum_{i=1}^{n}y_i\nabla\ell(y_i\theta_{t}^{T}x_i)z.
\end{equation}

By combining (\ref{transboundatt}) and (\ref{nablaLhatsimp}), we have:
\begin{equation}\label{boundattsimp}
\begin{aligned}
&\mathbb{E}\left[\hat{L}(\theta_{t+1})-\hat{L}(\theta_{t})\right] \\
&\leq\alpha G\frac{1}{n}\sum_{i=1}^{n}\mathbb{E}_{z}\left[-\nabla L(\theta_{t})+L\|\theta_{t}\|\|z\|^2\right] \\
&\quad +\alpha G\frac{1}{n}\sum_{i=1}^{n}\mathbb{E}_{z}\left[L\|\theta_{t}\|\|z\|^2-\frac{1}{n}\sum_{j=1}^{n}y_j\nabla\ell(y_j\theta_{t}^{T}x_j)z^2\right] \\
&\overset{(G)}{\leq}-\alpha G\frac{1}{n}\sum_{i=1}^{n}\nabla L(\theta_{t})+\alpha G(2L\|\theta_{t}\|+G)\mathbb{E}_{z}\left[\|z\|^2\right].
\end{aligned}
\end{equation}

For random variable X, we have:
\begin{equation}\label{EX2}
\mathbb{E}(X^2)=\mathbb{E}^2(X)+v(X),
\end{equation}
where $v(X)$ denotes the variance of X.

By (\ref{EX2}) and note that the random variable $z\sim(0,\sigma^2)$, (\ref{boundattsimp}) can be transferred to:
\begin{equation}\label{boundattfinal}
\begin{aligned}
&\mathbb{E}\left[\hat{L}(\theta_{t+1})-\hat{L}(\theta_{t})\right] \\
&\leq-\alpha G\frac{1}{n}\sum_{i=1}^{n}\nabla L(\theta_{t})+\alpha G(2L\|\theta_{t}\|+G)d\sigma^2 \\
&\overset{(G)}{\leq}G^2\alpha+\alpha G(2L\|\theta_{t}\|+G)d\sigma^2.
\end{aligned}
\end{equation}

By summing (\ref{boundattfinal}) over $T$ iterations and note that $\|\theta\|\leq D$:
\begin{equation}\label{boundT}
\mathbb{E}\left[\hat{L}(\theta_{T})-\hat{L}(\theta_{0})\right] \leq G^2\alpha T+\alpha G(2LD+G)d\sigma^2T.
\end{equation}

Then, considering the gap between $\hat{L}(\theta_{0})$ and $L^{*}$:
\begin{equation}\label{gap0*}
\begin{aligned}
\mathbb{E}\left[\hat{L}(\theta_{0})-L^{*}\right]&=\mathbb{E}\left[\frac{1}{n}\sum_{i=1}^{n}\ell(y_i\theta_{0}^{T}(x_i+z))-L^{*}\right] \\
&\overset{(G)}{\leq}\mathbb{E}\left[G\left|y_i\theta_{0}^{T}z\right|+(L(\theta_{0})-L^{*})\right] \\
&\leq G\|\theta_0\|\mathbb{E}\left[\|z\|\right]+(L(\theta_{0})-L^{*}) \\
&=L(\theta_{0})-L^{*}.
\end{aligned}
\end{equation}

If $\ell(\cdot)$ is $L$-smooth, we have:
\begin{equation}\label{Lsmooth}
L(\theta_{0})-L^{*}\leq\left<\nabla L(\theta^{*}),\theta_{0}-\theta^{*}\right>+\frac{L}{2}\|\theta_{0}-\theta^{*}\|^2,
\end{equation}
where $\theta^{*}$ denotes the optimal model and $\nabla L(\theta^{*})=0$.

Then, by (\ref{gap0*}) and (\ref{Lsmooth}), the inequality holds:
\begin{equation}\label{gap0*simp}
\mathbb{E}\left[\hat{L}(\theta_{0})-L^{*}\right]\overset{(L)}{\leq}\frac{L}{2}\|\theta_{0}-\theta^{*}\|^2.
\end{equation}

Then, by combination of (\ref{boundT}) and (\ref{gap0*simp}), we have:
\begin{equation}
\mathbb{E}\left[\hat{L}(\theta_{T})-L^{*}\right]\leq G^2\alpha T+\frac{L}{2}\|\theta_{0}-\theta^{*}\|^2+\alpha G(2LD+G)d\sigma^2T.
\end{equation}

Taking $\sigma$ the same as in (\ref{the1}), we have:
\begin{equation}
\mathbb{E}\left[\hat{L(\theta_{T})}-L^{*}\right]\leq O\left(\frac{\alpha(2LD+G)G^3d\log^2(n)\log(1/\delta)}{n(n-1)\sqrt{\Delta}\epsilon^2}\right),
\end{equation}
when $T=\widetilde{O}\left(\log(\frac{n(n-1)\sqrt{\Delta}\epsilon^2}{\alpha(2LD+G)G^3d\log(1/\delta)})\right)$, where the notation $\tilde{O}(\cdot)$ is similar to $O(\cdot)$, but hiding factors polynomial in $\log n$ and $\log(1/\delta)$.
\end{proof}

\subsection{Theorem 3}
\begin{proof}
Taking $M_t,P,Q,A,B,B',C$ the same as in A.1.

Note that the loss function $\ell(\theta,x,y)$ satisfies the Polyak-Lojasiewicz condition ($PL$), we have:
\begin{equation}
C \geq \frac{n-1}{n}\sqrt{2\mu(\ell(\theta_{t})-\ell^{*})}.
\end{equation}

As a result, in the moments accountant method:
\begin{equation}
\alpha_{M_t}(\lambda)\overset{(PL)}{\leq}\frac{2\lambda(\lambda+1)G^2}{\sqrt{2\mu(\ell(\theta_{T})-\ell^{*})}n(n-1)\sigma^2}.
\end{equation}

The factor $\sqrt{2\mu(\ell(\theta_{T})-\ell^{*})}$ can be seemed as a constant, then:
\begin{equation}
\alpha_{M_t}(\lambda) \leq c_1\frac{2\lambda(\lambda+1)G^2}{n(n-1)\sigma^2},
\end{equation}
for some constant $c_1$.

By summing $T$ iterations, for some constant $c_2$, we have:
\begin{equation}
\alpha_M(\lambda)\leq c_2\frac{\lambda^2G^2T}{\sigma^2n(n-1)}.
\end{equation}

Taking $\sigma$ the same as in (\ref{the1}), it can be guaranteed that:
\begin{equation}
\alpha_M(\lambda) \leq \frac{\lambda\epsilon}{2},
\end{equation}
and as a result:
\begin{equation}
\delta \leq \exp(\frac{-\lambda\epsilon}{2}),
\end{equation}
for some constant $c$, which means ($\epsilon,\delta$)-differential privacy due to Theorem 2.2 in \cite{15}.
\end{proof}

\end{document}